\documentclass[final,12pt]{elsarticle}

\usepackage{amssymb}
\usepackage{amsmath,amsfonts}
\usepackage{booktabs}
\usepackage{multirow}
\usepackage{siunitx}
\usepackage{subfig}
\usepackage[breaklinks=true]{hyperref}
\usepackage{todonotes}
\usepackage{arydshln}
\usepackage{graphbox}
\usepackage{textpos}

\makeatletter
\def\ps@pprintTitle{   \let\@oddhead\@empty
   \let\@evenhead\@empty
   \def\@oddfoot{\reset@font\hfil\thepage\hfil}
   \let\@evenfoot\@oddfoot
}
\makeatother

\newcommand{\tabtt}[1]{\texttt{\footnotesize{}#1}}

\journal{Knowledge-Based Systems}

\begin{document}

\begin{frontmatter}

\title{Generating Multidimensional Clusters With Support Lines}

\author{Nuno Fachada}
\ead{nuno.fachada@ulusofona.pt}

\author{Diogo de Andrade}
\ead{diogo.andrade@ulusofona.pt}

\address{Lusófona University, COPELABS\\
Campo Grande, 376, Lisbon, Portugal}

\begin{abstract}

Synthetic data is essential for assessing clustering techniques,
complementing and extending real data, and allowing for more complete
coverage of a given problem's space. In turn, synthetic data generators
have the potential of creating vast amounts of data---a crucial activity
when real-world data is at premium---while providing a well-understood
generation procedure and an interpretable instrument for methodically
investigating cluster analysis algorithms. Here, we present \textit{Clugen},
a modular procedure for synthetic data generation, capable of creating
multidimensional clusters supported by line segments using arbitrary
distributions. \textit{Clugen} is open source, comprehensively unit tested and
documented, and is available for the Python, R, Julia, and MATLAB/Octave
ecosystems. We demonstrate that our proposal can produce rich and
varied results in various dimensions, is fit for use in the assessment of
clustering algorithms, and has the potential to be a widely used framework
in diverse clustering-related research tasks.

\end{abstract}


\end{frontmatter}

\begin{textblock*}{190mm}(-3cm,-19cm)
    \noindent \footnotesize The peer-reviewed version of this paper is
    published in Knowledge-Based Systems at
    \url{https://doi.org/10.1016/j.knosys.2023.110836}.
    This version is typeset by the authors and differs only in pagination and
    typographical detail.
\end{textblock*}

\section{Introduction}
\label{sec:intro}

Synthetic data generation is an important step in the evaluation and improvement
of clustering and outlier detection techniques \cite{li2022ac,korzeniewski2014empirical}.
It allows for the creation of cluster structures that resemble real-world datasets,
have specific characteristics---for example, focusing on rare and edge cases---and/or
are representative of a desired problem space \cite{shand2021hawks,smith2015generating}.
Synthetic data generators can potentially create limitless amounts of data, which
is particularly useful when real-world data is scarce or difficult to obtain
\cite{pei2006synthetic,fachada2020generatedata}. Real data can also have high cost,
unbalanced quality, fail to adequately cover the problem space, and generally be
poorly understood \cite{li2022ac,shand2019evolving,macia2014towards}. Furthermore,
real data may lack correct class membership, making it unsuitable for objective
assessments of clustering performance using external cluster validation indices
\cite{shand2021hawks}. While real-world datasets and scenarios are the definitive
test for assessing clustering techniques \cite{iglesias2019mdcgen}, synthetic data
generation remains a relevant---if not crucial---tool for this purpose, avoiding
many of the problems associated with real data \cite{pei2006synthetic}.

Data created with an open source and properly documented synthetic data
generator has, by definition, a known generation process, and therefore
provides the cluster memberships, as well as the assumptions implicit in the
generating procedure.
Consequently, this type of data enables the identification
of persistent biases, merits, and limitations of clustering methods,
promoting neutral and objective performance evaluation, while highlighting the
underlying factors responsible for such performance
\cite{shand2021hawks,shand2019evolving,qiu2006generation}. Different algorithms
can perform in a significantly different manner with the same input data, being
affected by factors such as number of clusters, degree of cluster separation,
noise, and/or cluster shape (in particular, cluster elongation)
\cite{pei2006synthetic,qiu2006generation,melnykov2012mixsim,handl2006multi}.
Since there is no best algorithm to use in all cases, synthetic data generation
can help pinpoint how a given technique performs in a number of different,
user-defined scenarios. In sum, data generation techniques are an interpretable
vehicle for systematically investigating the robustness and performance of
algorithms in various areas of cluster analysis
\cite{melnykov2012mixsim,steinley2005oclus}.

In this paper we present \textit{Clugen}, a synthetic data generation algorithm,
together with four open source, fully tested and documented software
implementations available as a library of functions for the Python
\cite{vanrossum2022python}, R \cite{r2022}, Julia \cite{bezanson2017julia},
and MATLAB/Octave \cite{matlab2022,octave2022}
ecosystems. The algorithm is able to generate multidimensional synthetic datasets
for testing and benchmarking clustering techniques and works by creating clusters
``around'' (supported by) line segments. The aim is to facilitate the production of
elongated clusters, similar to the work of Handl and Knowles \cite{handl2005cluster},
but achieving controllable cluster eccentricity by using line segments.
\textit{Clugen} is based on \textit{generateData}, an algorithm that also uses
support lines to create clusters, but is limited to 2D and a few predefined
distributions, and is only available for MATLAB/Octave
\cite{fachada2020generatedata}. Conversely, \textit{Clugen}  works
in multiple dimensions and allows the use of arbitrary distributions for cluster
generation. It is considerably efficient due to mainly using vectorized code while
avoiding iterative and conditional generation procedures (which may throw away or
reshape data that does not fit some requirement). Parameterization is simple by
default, although fully customizable, and therefore capable of producing a rich
set of characteristics for manipulating clustering performance.
More than providing a fixed algorithm, \textit{Clugen}'s implementations are data
generation toolboxes. Their components are provided as separate functions
which can be mixed, matched, and/or replaced by users. More generally, the goal is
to simplify---or at least allow---the creation of large amounts of synthetic data
with similar characteristics, possibly mimicking and complementing real data, e.g.,
for the purpose of determining which clustering algorithm is generally more adequate
for data with particular features, or for specific types of real, but scarce, data.

This paper is organized as follows. In Section~\ref{sec:background}, we review
related work on synthetic cluster generation and discuss how \textit{Clugen}
offers a novel and robust approach. The methods are presented in
Section~\ref{sec:methods}, starting with a detailed, step-by-step description of
the \textit{Clugen} algorithm (\ref{sec:methods:alg}), continuing with a number
of example parameterizations in higher dimensions (\ref{sec:methods:examples}),
and finalizing with a characterization of its four software implementations
(\ref{sec:methods:impl}). A concrete usage example, in the form of an experiment
in assessing clustering algorithms, is demonstrated in Section~\ref{sec:useexample}.
A discussion about \textit{Clugen}'s potential, as well as its limitations, takes
place in Section~\ref{sec:disc}. The paper closes with Section~\ref{sec:conclusions},
in which we offer some conclusions.

\section{Related Work}
\label{sec:background}

Several clustering-oriented synthetic data generators using various approaches
and methodologies have been proposed. Some of these were accompanied by a
software implementation, promoting their dissemination and use by other
researchers.

One of the earliest and most well-known procedures was proposed by Milligan
\cite{milligan1985algorithm}, where clusters are drawn from multivariate
normal distributions through diagonal covariance matrices, with points falling
outside 1.5 standard deviations of the mean on each dimension of the cluster
being rejected. The method allows the inclusion of outliers and random noise
and can generate data in 4, 6, or 8 dimensions with up to 5 clusters,
while guaranteeing that clusters do not overlap in the first dimension. The
method offers limited control over the number of clusters, number of dimensions,
cluster sizes, outliers, and noise. A third-party GPLv2-licensed C++
implementation is available on SourceForge \cite{pape2000clusutils}, but it does
not compile on modern systems without modification, and it is unclear how
accurately it reflects Milligan's algorithm.

\textit{SynDECA} is a tool for generating datasets to evaluate clustering
algorithms, enabling users to specify the total number of points, number of
clusters, and number of dimensions, randomly determining other structural
properties of the data, such as clusters' radius and percentage of noise
\cite{vennam2005syndeca}. The algorithm produces non-overlapping clusters in
various shapes (circle, ellipse, rectangle, square and irregular---the latter
being grown by ``sprinkling'' points along a random path). Clusters may be
rejected (and thus re-randomized) if they overlap. The same occurs for points
(or noise) if they fall outside (or inside) the clusters' radius. The user has
limited control over the generated datasets, and these tend to have an
artificial or handcrafted look. The authors provide a basic but functional C++
implementation\footnote{\url{https://sites.google.com/site/syndeca/}},
although no tests are provided and no license is specified.

Two generators, proposed by Handl and Knowles \cite{handl2005cluster}, were
developed with a focus on generating large numbers of clusters in high-dimensional
space. The first algorithm iteratively generates multivariate normal clusters by
stochastically building symmetric, positive definite covariance matrices,
rejecting clusters that overlap with existing ones. While this approach generally
creates elongated clusters when the number of dimensions is relatively low, clusters
tend to become spheroidal as the number of dimensions increases. The second
algorithm addresses this issue by creating ellipsoidal clusters with a main axis
in a random orientation, placing points at a normally distributed distance from
a uniformly random point on the major axis (points are rejected if placed
outside the ellipsoid's boundaries). A genetic algorithm is used to move cluster
origins with the goal of minimizing cluster overlap and overall variance in the
dataset. A C++ implementation is publicly
available\footnote{\url{https://personalpages.manchester.ac.uk/staff/Julia.Handl/generators.html}}
under a GPLv2 license, but does not compile on modern systems without
modification.

Another generator, \textit{OCLUS}, uses an analytic approach for creating
clusters with considerable overlap control \cite{steinley2005oclus}. Clusters
are drawn independently in each dimension from the uniform, normal, gamma, or
triangular distribution. Among several input parameters, \textit{OCLUS}
requires the specification of overlaps in each pair of adjacent clusters for
all dimensions. While this approach allows for substantial overlap control,
the proposed technique is unable to generate clusters with known skew and
kurtosis. \textit{OCLUS} is implemented in MATLAB, but the code must be
requested from the authors and therefore is not readily available.

Qiu and Joe \cite{qiu2006generation} proposed an improvement on Milligan's
original algorithm by determining the distance between clusters through a
user-provided ``separation index'' value. In practice, the method uses a
geometric framework to ensure a minimum degree of separation between the closest
neighboring clusters in the first dimension, with no guarantees made for the
remaining dimensions. To achieve this, the covariance matrices are iteratively
scaled until this minimum separation is satisfied. Like Milligan's algorithm,
the method allows adding noise and outliers to the generated dataset and is
likewise limited to generating clusters from the multivariate normal
distribution. The method is publicly available as a GPLv2-licensed R
package\footnote{\url{https://cran.r-project.org/package=clusterGeneration}},
although at time of writing, collaborative source control does not exist and
no tests are included in the source package.

Pei and Za{\"\i}ane proposed a distribution and transformation-based technique
for 2D synthetic data generation \cite{pei2006synthetic}. Clusters are drawn
from the standard uniform or normal distributions---or from several predefined
shapes---and are manipulated through various mathematical transformations. Users
can specify the (clustering) difficulty level, number of points, number of
clusters, several cluster properties (percentage of points, density, shape,
location), and the outlier level. Increasing difficulty levels yield clusters
with more complex shapes while increasing outlier levels produce outliers with
clearer patterns. The method is implemented in Java and has a graphical user
interface for manipulating the parameters and displaying the generated datasets
for visual inspection. However, there is no clear way of obtaining the software.

The widely used \textit{scikit-learn} Python machine learning framework
\cite{pedregosa2011scikit} also includes several clustering-oriented synthetic
data generators. The one most directly comparable to the generators discussed
here---although substantially more limited---is termed \textit{blobs}, and
generates isotropic normal clusters, allowing the user to specify the number of
dimensions, number of clusters, cluster centers, and standard deviation (equal
for all clusters). Other generators in this framework are 2D-only and generate
very specific shapes. Nonetheless, \textit{scikit-learn} is widely available,
well-tested and documented, and offered under a liberal BSD-3-clause license.

\textit{MixSim} is an R package (with a kernel written in C) for simulating
mixtures of normal distributions---drawn from stochastically built covariance
matrices---with different levels of overlap between mixture components
\cite{melnykov2012mixsim}. Pairwise overlap, a measure proposed by the same
authors in a previous publication \cite{maitra2010simulating}, is used to
compute the overlap between mixture components. The tool also allows the
simulation of outliers and noise and offers functionality for determining
several indices for measuring the classification accuracy between two
partitionings. \textit{MixSim} is
available\footnote{\url{https://cran.r-project.org/package=MixSim}} under the
GPLv2 or GPLv3 licenses. The source package contains some demos and examples,
but no explicit tests, and does not appear to be source-controlled or provide
collaboration channels.

\textit{ELKI} is an unsupervised learning framework that, in a similar fashion to
\textit{scikit-learn}, offers a synthetic cluster generator
\cite{schubert2015framework}. This generator, however, is considerably more powerful,
being able to generate multidimensional clusters drawn from the uniform, normal, and
gamma distributions, dimension by dimension---not unlike \textit{OCLUS}. The
generator also allows controlling cluster sizes, as well as manipulating clusters
through rotations, translations, scaling, and clipping. Each created dataset has to
be specifically and thoroughly described in XML, making \textit{ELKI}'s generator
inappropriate for high throughput synthetic data generation. \textit{ELKI} is
written in Java and is distributed under the AGPLv3 license. The framework is
available on GitHub \footnote{\url{https://elki-project.github.io/}}---the
repository seems quite active---and the code appears to have basic testing and
a continuous integration (CI) setup.

In turn, \textit{MDCGen} is a multidimensional, highly-parameterizable, and
clustering-oriented synthetic data generator \cite{iglesias2019mdcgen}. Several
distributions are supported out of the box---together with custom, user-specified
distributions---which can be combined dimension by dimension or by defining
overall intra-cluster distances. The generator provides, via a grid-based scheme,
some degree of control over cluster separation/overlap, as well as inclusion of
outliers and noise, feature correlation, cluster rotation, and creation of
subspace clusters (i.e., clusters clearly detectable only in some dimensions).
Parameterization is performed through a JSON file and can be somewhat complex
and unintuitive. However, all parameters are optional, making \textit{MDCGen} much
simpler to start with than, for example, \textit{ELKI}'s generator. The tool is
available for both MATLAB\footnote{\url{https://github.com/CN-TU/mdcgen-matlab}}
and Python\footnote{\url{https://github.com/CN-TU/mdcgenpy}} under the GPLv3
license. The code is source-controlled and available on GitHub. While the MATLAB
version works as stated (for MATLAB $\ge$ R2017a, not R2016a as mentioned in the
README file) and includes some tests (but no CI workflows), the Python
implementation is not usable as documented on modern Python ($\ge$ 3.8) and
does not appear to be unit tested.

With the purpose of identifying dataset characteristics relevant for
differentiating the performance of clustering algorithms, as well as of evolving
benchmarks with specific characteristics, the \textit{HAWKS} generator uses an
evolutionary algorithm (EA) to optimize datasets according to a given fitness
function, employing constraints such as local cluster overlap and cluster elongation
\cite{shand2021hawks,shand2019evolving}. The authors experimented with two
optimization modes. The first optimizes the datasets towards a user-specified
value of an internal cluster validity index---the authors experimented with the
silhouette width \cite{rousseeuw1987silhouettes}---to control the general
difficulty of the datasets. The second evolves datasets for maximizing the
performance difference between two clustering algorithms to highlight the
relative strengths and weaknesses of each. The initial EA
population is composed of datasets with multivariate normal clusters generated
from valid covariance matrices, possibly rotated and scaled, in a similar
fashion to Qiu and Joe's generator \cite{qiu2006generation}. The user can define
the number of clusters and total number of points, which can be randomly
assigned or evenly distributed to the existing clusters (either way adding up to
the specified total number of points). Cluster sizes remain fixed across all
individuals during evolution so that the EA can focus only on optimizing the
distribution parameters. The initial means (cluster centers) are sampled from a
multidimensional uniform distribution using a user-defined interval. The authors
state that \textit{HAWKS} can create diverse datasets able to cause more
variation in the performance of clustering algorithms. Due to the use of an EA,
the generator is considerably slow, which may limit its use as a high-throughput
solution. The code is available \footnote{\url{https://github.com/sea-shunned/hawks}}
under an MIT license, is easily installable and reasonably well-tested (although
no CI workflows are set up). At the time of writing, the repository sees little
activity and the code uses deprecated features on one of its supporting libraries,
which may be a cause of concern when using this generator in the long term.

The \textit{generateData} procedure produces 2D data clusters along line segments
\cite{fachada2020generatedata}. The goal was to mimic real data often observed
after dimensionality reduction of spectrometric measurements and was originally
employed to augment existing but quantity-limited datasets of this kind
\cite{fachada2014spectrometric}. Besides the number of points and clusters, the
user can control the average direction of the cluster-supporting lines,
how parallel the lines are to each other, and the distribution parameters of
cluster points along and around the lines. This method is the precursor of the
\textit{Clugen} algorithm presented in this paper, and although
limited---especially in terms of dimensionality---it has been employed in
several studies, as will be discussed in Section~\ref{sec:disc}. The method is
offered as a MATLAB/Octave
function\footnote{\url{https://github.com/nunofachada/generateData}} under the
MIT license, though no formal tests are included in its repository.

The \textit{AC} data generator uses the normal distribution to generate
multidimensional synthetic data, controlling the clusters' shape by building
them using a hierarchical approach, i.e., clusters can be composed of several
normally distributed sub-clusters \cite{li2022ac}. Leveraging this approach,
\textit{AC} supports Bezier curve-based clusters containing 200 subclusters,
for which users can optionally specify control points, versus these being
randomly defined. Clusters can also be grouped into layers, where each layer
retains control over the characteristics of their respective clusters. Clusters
can be placed using relative or absolute positioning, with the authors stating
that sample size, distribution size, cluster overlap, and shape are thoroughly
customizable. The generator allows editing previously created datasets in order
to add new clusters. \textit{AC} is oriented towards generating a large number
of datasets with similar characteristics in a high-throughput fashion. The
software appears to be written in a Java-based language, and while the authors
provide an executable .jar file\footnote{\url{http://ac.fwgenetics.org/}}---as
well as a web-based implementation---the source code is not available for
inspection. Consequently, no statements can be made about version control and
unit testing.

A very recent Python-based tool, \textit{repliclust}, proposes an
archetype-based interface for defining high-level geometric dataset
descriptions (through max-min parameters), from which users can create datasets
with similar cluster characteristics \cite{zellinger2023repliclust}. The tool
generates ellipsoidal clusters with various distributions, sampling points
around its principal axes, although actual cluster orientations are not
user-controllable.
A level of cluster overlap
control is provided by leveraging the error rate of the best linear classifier
between pairwise clusters, working best when clusters are normally distributed;
however, even using some approximations for this purpose, the algorithm
is considerably inefficient when combining many dimensions, clusters and/or
points. The code\footnote{\url{https://github.com/mzelling/repliclust}},
provided under a BSD-3-clause license, is modular and allows users to change and
control parts of the algorithm and subsequent data generation. It has good test
coverage and quality documentation is available, contrary to most of the tools
described so far. Nonetheless, there are packaging issues at time of writing,
and the library will only work after the user manually installs the required
dependencies.

The generators discussed thus far either focus on refining very specific
cluster characteristics or use relatively naive approaches to data generation
\cite{shand2019evolving}. For example, many focus on creating multivariate
normal clusters and then manipulating or composing them into the desired shapes,
or to (mostly) assure that certain specific cluster overlap and/or separation is
obtained at the cost of some flexibility and performance. Other generators,
such as \textit{HAWKS} or \textit{ELKI}, are not
appropriate for high-throughput data generation, the first being too slow due
to using EAs for optimizing the desired features, while the latter requires very
detailed parameterization for each dataset produced. Yet others are simply not
publicly available, do not work out of the box, or fail to follow software
development best practices, such as including adequate documentation, thorough
test coverage, or implementing a CI pipeline. \textit{Clugen} aims to bridge this
gap, providing not only an algorithm for spatially intuitive generation of
clustering-oriented datasets aimed at mimicking real data but also a
meta-algorithmic framework, allowing users to customize output, possibly using
methods available in other generators. Further, with four open source, fully
tested, and CI-enabled software implementations---available in GitHub and the
respective language package distribution hubs---\textit{Clugen} goes one step
further to avoid bit rot and achieve long-term sustainability.

\section{Methods}
\label{sec:methods}

In this section we describe \textit{Clugen}, beginning with a comprehensive,
step-by-step characterization of the algorithm with illustrative cases
(Subsection~\ref{sec:methods:alg}), presenting several complementing examples
in Subsection~\ref{sec:methods:examples}, and finalizing with a depiction of its
four software implementations (Subsection~\ref{sec:methods:impl}). These
implementations are equivalent and expose the algorithm via the \texttt{clugen()}
function, which receives various mandatory and optional parameters, listed in
Table~\ref{tab:paramsmand} and Table~\ref{tab:paramsopt}, respectively. These
tables present both the symbol and the name of the parameter. The former will be
used when describing algorithm steps in a formal mathematical notation, while the
latter represents the name of the parameter as accepted by the \texttt{clugen()}
function. This function---as well as the several functions denoted in typewriter
font---are included in the four \textit{Clugen} implementations and accept the
same parameters (and return the same values) across implementations. These
functions are introduced in context during the algorithm's description in
Subsection~\ref{sec:methods:alg} and are summarized in
Subsection~\ref{sec:methods:impl}.

\begin{table}[]
    \caption{Mandatory parameters of the \texttt{clugen()} function.}
    \label{tab:paramsmand}
    {\small
    \begin{tabular}{llp{10cm}}
    \toprule
    \multicolumn{2}{l}{Symbol / Name} & Description \\
    \midrule
    $n$
        & \tabtt{num\_dims}
        & Number of dimensions.\\
    $c$
        & \tabtt{num\_clusters}
        & Number of clusters.\\
    $p$
        & \tabtt{num\_points}
        & Total number of points to generate.\\
    $\mathbf{d}$
        & \tabtt{direction}
        & Average direction of cluster-supporting lines (vector of length $n$).\\
    $\theta_\sigma$
        & \tabtt{angle\_disp}
        & Angle dispersion of cluster-supporting lines (radians).\\
    $\mathbf{s}$
        & \tabtt{cluster\_sep}
        & Average cluster separation in each dimension (vector of length $n$).\\
    $l$
        & \tabtt{llength}
        & Average length of cluster-supporting lines.\\
    $l_\sigma$
        & \tabtt{llength\_disp}
        & Length dispersion of cluster-supporting lines.\\
    $f_\sigma$
        & \tabtt{lateral\_disp}
        & Cluster lateral dispersion, i.e., dispersion of points from their
          projections on the cluster-supporting line.\\

    \bottomrule
    \end{tabular}
    }
\end{table}

\begin{table}[]
    \caption{Optional parameters of the \texttt{clugen()} function.}
    \label{tab:paramsopt}
    {\small
    \begin{tabular}{lllp{6cm}}
    \toprule
    \multicolumn{2}{l}{Symbol / Name} & Default & Description \\
    \midrule
    $\phi$
        & \tabtt{allow\_empty}
        & \textit{False}
        & Allow empty clusters? \\
    $\mathbf{o}$
        & \tabtt{cluster\_offset}
        & $(0, \dots, 0)$
        & Offset to add to all cluster centers (vector of length $n$).\\
    $p_\text{proj}()$
        & \tabtt{proj\_dist\_fn}
        & \texttt{"norm"}
        & Distribution of point projections along cluster-supporting lines.\\
    $p_\text{final}()$
        & \tabtt{point\_dist\_fn}
        & \texttt{"n-1"}
        & Distribution of final points from their projections.\\
    $c_s()$
        & \tabtt{clusizes\_fn}
        & \tabtt{clusizes()}
        & Distribution of cluster sizes.\\
    $c_c()$
        & \tabtt{clucenters\_fn}
        & \tabtt{clucenters()}
        & Distribution of cluster centers.\\
    $l()$
        & \tabtt{llengths\_fn}
        & \tabtt{llengths()}
        & Distribution of line lengths.\\
    $\theta_\Delta()$
        & \tabtt{angle\_deltas\_fn}
        & \tabtt{angle\_deltas()}
        & Distribution of line angle deltas (w.r.t. $\mathbf{d}$).\\
    \bottomrule
    \end{tabular}
    }
\end{table}

\subsection{The \textit{Clugen} Algorithm}
\label{sec:methods:alg}

The \textit{Clugen} algorithm generates multidimensional data by using line
segments to support the different clusters. The position, orientation, and
length of each line segment guide where points of the corresponding cluster are
placed. For brevity, \emph{line segments} will be referred to as \emph{lines}
for the remainder of this paper.

\begin{figure}[!t]
    \centering

    \includegraphics[width=0.991\linewidth]{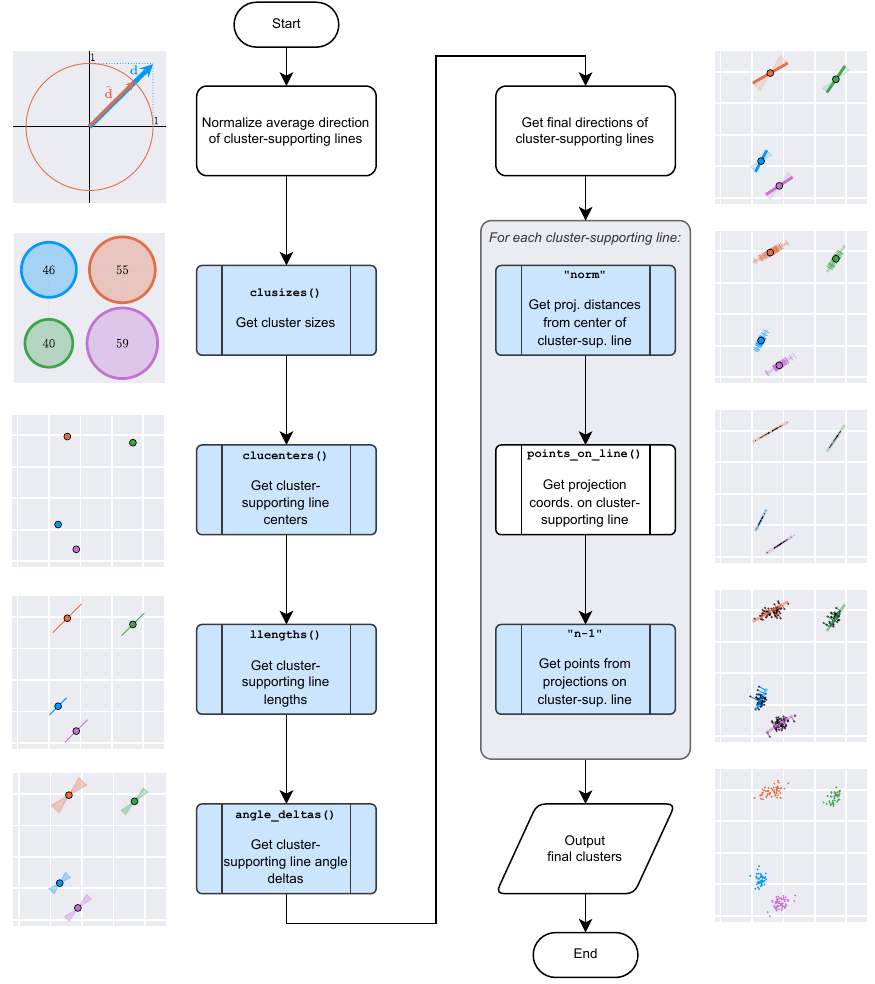}

    \caption{Flowchart of the \textit{Clugen} algorithm. Functions in light
    blue are stochastic and can be swapped by the user through the optional
    parameters (Table~\ref{tab:paramsopt}). The
    \texttt{"norm"} and \texttt{"n-1"} strings are aliases for
    the functionality described in Sections~\ref{sec:methods:alg:projs} and
    \ref{sec:methods:alg:points}, respectively.
    A stylized 2D example of the algorithm's steps is shown on respective side
    images. The example was generated with Julia's \textit{Clugen}
    implementation and mandatory parameters (Table~\ref{tab:paramsmand}) set to
    $n=2$, $c=4$, $p=200$, $\mathbf{d}=(1,1)$, $\theta_\sigma=\pi/16$,
    $\mathbf{s}=(10,10)$, $l=10$, $l_\sigma=1.5$, and $f_\sigma=1$. Optional
    parameters were left to their defaults.
    }
    \label{fig:flow}
\end{figure}

Given the mandatory parameters (Table~\ref{tab:paramsmand}), as well as the
optional ones (Table~\ref{tab:paramsopt}), the algorithm generates data
according to the flowchart and stylized example presented in
Fig.~\ref{fig:flow}. The functions presented in light blue can be provided by
the user as optional parameters, as shown in Table~\ref{tab:paramsopt}. However,
the algorithm specifies useful defaults---also presented in
Table~\ref{tab:paramsopt}. The most important and/or obvious mandatory
parameters are the number of dimensions, $n$, the number of clusters, $c$, the
total number of points, $p$,
and the average direction
of cluster-supporting lines, $\mathbf{d}$. In the context of the mandatory
parameters, ``average'' and ``dispersion'' refer to measures of central
tendency and variability, respectively, although their concrete meaning depends
on the statistical distributions defined in the optional parameters shown in
Table~\ref{tab:paramsopt}, and highlighted in light blue in Fig.~\ref{fig:flow}.

Subsections~\ref{sec:methods:alg:norm}--\ref{sec:methods:alg:final} detail the
algorithm's steps, relating and contextualizing the various mandatory and
optional parameters, while describing the default values and distributions for
the latter.
While \textit{Clugen} is based on the idea of cluster-supporting lines
introduced in \textit{generateData}, the latter's simple trigonometry-based
approach is not generalizable to multiple dimensions. Therefore, the following
steps were developed from the ground up using principles from fundamental
geometry with the goal of modularizing and extending the concept of
cluster-supporting lines to multiple dimensions, while allowing arbitrary
distributions at each stochastic step---which default to
\textit{generateData}-like fixed presets.

\subsubsection{Normalize the average direction of cluster-supporting lines}
\label{sec:methods:alg:norm}

The first step of the \textit{Clugen} algorithm is to convert the average
direction of the cluster-supporting lines, $\mathbf{d}$, to a unit vector,
which is required for later calculations. More specifically:

\begin{equation}
    \hat{\mathbf{d}} = \cfrac{\mathbf{d}}{\left\lVert\mathbf{d}\right\rVert}
\end{equation}

\noindent where $\hat{\mathbf{d}}$ is the normalized version of $\mathbf{d}$.
This step simplifies the use of the algorithm, allowing users to provide any
direction to the \texttt{clugen()} function, and not necessarily a normalized
one. Note that $\hat{\mathbf{d}}$ and $\mathbf{d}$ are  $n$-dimensional
direction vectors, where $n$ is the dimensionality of the generated data.

\subsubsection{Determine cluster sizes}
\label{sec:methods:alg:clusizes}

Cluster sizes are given by the function referenced in $c_s()$ according to:

\begin{equation}
    \mathbf{c_s} = c_s(c, p, \phi)
\end{equation}

\noindent where $\mathbf{c_s}$ is a $c$-dimensional integer vector containing
the final cluster sizes, $c$ is the number of clusters, $p$ is (a hint for) the
total number of points to generate, and $\phi$ is a boolean that determines
whether empty clusters are acceptable.

The $c_s()$ function reference is an optional parameter, allowing users to
customize the cluster sizing behavior. By default, $c_s()$ references the
\texttt{clusizes()} function, which behaves according to the following
algorithm:

\begin{enumerate}
    \item Determine the size $p_i$ of each cluster $i$ according to:
        \begin{equation*}
            p_i\sim\left\lfloor\max\left(\mathcal{N}(\frac{p}{c}, (\frac{p}{3c})^2),0\right)\right\rceil
        \end{equation*}
        \noindent where $\lfloor\rceil$ denotes the round to nearest integer
        function, and $\mathcal{N}(\mu,\sigma^2)$ represents the normal
        distribution with mean $\mu$ and variance $\sigma^2$.
    \item Assure that the final cluster sizes add up to $p$ by incrementing the
        smallest cluster size while $\sum_{i=1}^c p_i<p$ or decrementing the
        largest cluster size while $\sum_{i=1}^c p_i>p$. This step is delegated
        to the \texttt{fix\_num\_points()} helper function.
    \item If $\neg\phi\wedge p\ge c$ then, for each empty cluster $i$ (i.e.,
        $p_i=0$), increment $p_i$ and decrement $p_j$, where $j$ denotes the
        largest cluster. This step is delegated to the \texttt{fix\_empty()}
        helper function.
\end{enumerate}

\begin{figure}[]
    \centering

    \subfloat[\label{fig:clusizes:normal}]{
        \includegraphics[width=0.245\linewidth]{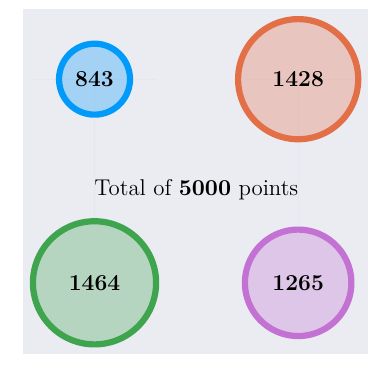}}
    \subfloat[\label{fig:clusizes:unif}]{
        \includegraphics[width=0.245\linewidth]{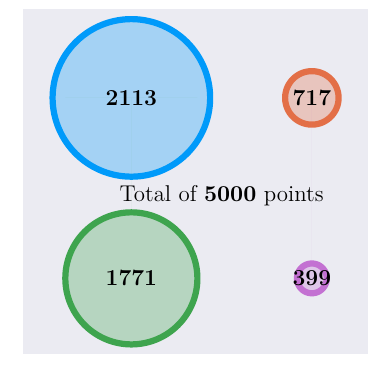}}
    \subfloat[\label{fig:clusizes:poisson}]{
        \includegraphics[width=0.245\linewidth]{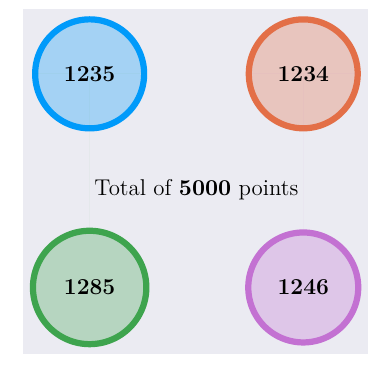}}
    \subfloat[\label{fig:clusizes:poisson_notpfix}]{
        \includegraphics[width=0.245\linewidth]{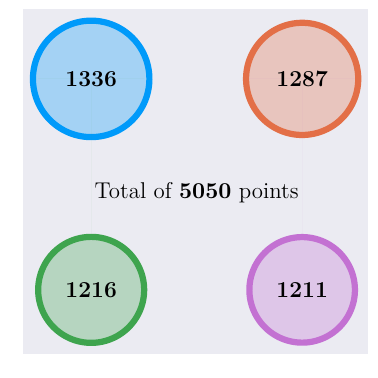}}

    \caption{ Possible cluster sizes with $c=4$ and $p=5000$, and $c_s()$ set as
        follows ((b)--(d) are custom user functions):
        (a) normal distribution (discretized) with total points correction
        (the default, implemented by the \texttt{clusizes()} function);
        (b) discrete uniform distribution with total points correction;
        (c) Poisson distribution with total points correction;
        (d) Poisson distribution, no correction.}
    \label{fig:clusizes}
\end{figure}

Fig.~\ref{fig:clusizes} demonstrates possible cluster sizes with various
definitions of $c_s()$ for $c=4$ and $p=5000$. The default behavior, implemented
in the \texttt{clusizes()} function, is shown in Fig.~\ref{fig:clusizes:normal},
while Figs.~\ref{fig:clusizes:unif}--\ref{fig:clusizes:poisson_notpfix} present
results obtained with custom user functions. Fig.~\ref{fig:clusizes:unif}
displays cluster sizes obtained with the discrete uniform distribution over
$\left\{1, 2, \ldots, \frac{2p}{c}\right\}$, corrected with
\texttt{fix\_num\_points()}. In turn, Fig.~\ref{fig:clusizes:poisson} highlights
cluster sizes obtained with the Poisson distribution with $\lambda=\frac{p}{c}$,
also corrected with \texttt{fix\_num\_points()}. The cluster sizes shown in
Fig.~\ref{fig:clusizes:poisson_notpfix} were determined with the same
distribution (Poisson, $\lambda=\frac{p}{c}$), but were not corrected. Thus,
cluster sizes do not add up to $p$, highlighting the fact that this is not a
requirement of the \textit{Clugen} algorithm, i.e., user-defined $c_s()$
implementations can consider $p$ a hint rather than an obligation. Likewise,
they can ignore $\phi$ (the \texttt{allow\_empty} parameter), although that is
probably not advisable, as it can lead the user in error. For the remainder of
this paper, we will reference the effective final total number of points as
$p^\star$, which may or may not be equal to $p$.

\subsubsection{Determine cluster centers}
\label{sec:methods:alg:centers}

Cluster centers are given by the function referenced in $c_c()$ according to:

\begin{equation}
    \mathbf{C} = c_c(c, \mathbf{s}, \mathbf{o})
\end{equation}

\noindent where $\mathbf{C}$ is a $c \times n$ matrix containing the final
cluster centers, $c$ is the number of clusters, $\mathbf{s}$ is the average
cluster separation ($n \times 1$ vector), and $\mathbf{o}$ is an $n \times 1$
vector of cluster offsets.

The $c_c()$ function reference is an optional parameter, allowing users to
customize how cluster centers are determined. By default, $c_c()$ references the
\texttt{clucenters()} function, which determines the cluster centers according
to:

\begin{equation}
    \mathbf{C}=c\mathbf{U} \cdot \operatorname{diag}(\mathbf{s}) + \mathbf{1}\,\mathbf{o}^T
    \label{eq:cc}
\end{equation}

\noindent where $\mathbf{U}$ is a $c \times n$ matrix of random values drawn
from the uniform distribution between -0.5 and 0.5, and $\mathbf{1}$ is a
$c \times 1$ vector with all entries equal to 1.

\begin{figure}[]
    \centering

    \subfloat[\label{fig:clucenters:unif}]{
        \includegraphics[width=0.245\linewidth]{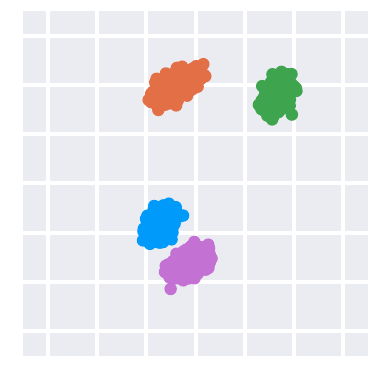}}
    \subfloat[\label{fig:clucenters:hand}]{
        \includegraphics[width=0.245\linewidth]{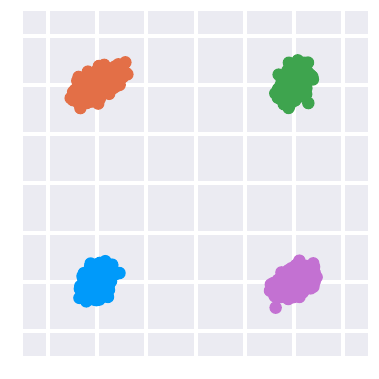}}

    \caption{The output of the \textit{Clugen} algorithm for two different
    $c_c()$-referenced functions for finding cluster centers: a) the
    default, using the uniform distribution; b) hand-picked centers. Remaining
    parameters are the same as in Fig.~\ref{fig:flow}, except for $p$, which is
    set to \num{5000}.}
    \label{fig:clucenters}
\end{figure}

Fig.~\ref{fig:clucenters} shows scatter plots of the results generated by the
\textit{Clugen} algorithm for two different definitions of $c_c()$, namely using
the uniform distribution (the default, as described by Eq.~\ref{eq:cc},
Fig.~\ref{fig:clucenters:unif}), and direct specification of cluster
centers with a custom user function (Fig.~\ref{fig:clucenters:hand}).

\subsubsection{Determine lengths of cluster-supporting lines}
\label{sec:methods:alg:lengths}

The lengths of the cluster-supporting lines are given by the function referenced
in $l()$ according to:

\begin{equation}
    \pmb{\ell} = l(c, l, l_\sigma)
\end{equation}

\noindent where $\pmb{\ell}$ is a $c$-dimensional vector containing the final
lengths of the cluster-supporting lines, $c$ is the number of clusters, $l$ is
the average length, and $l_\sigma$ is the length dispersion.

\begin{figure}[]
    \centering

    \subfloat[\label{fig:llengths:fnormal}]{
        \includegraphics[width=0.245\linewidth]{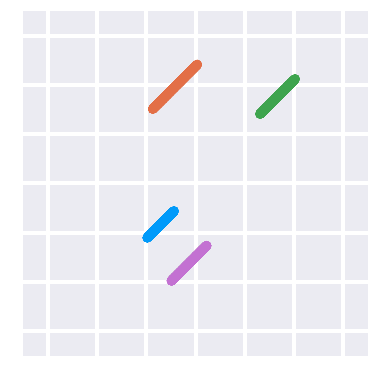}}
    \subfloat[\label{fig:llengths:poisson}]{
        \includegraphics[width=0.245\linewidth]{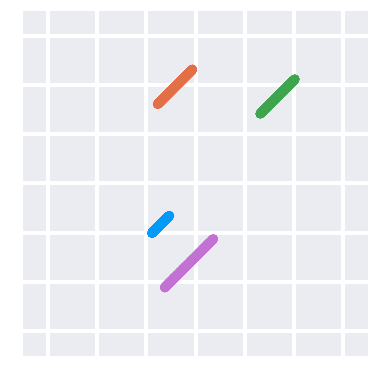}}
    \subfloat[\label{fig:llengths:unif}]{
        \includegraphics[width=0.245\linewidth]{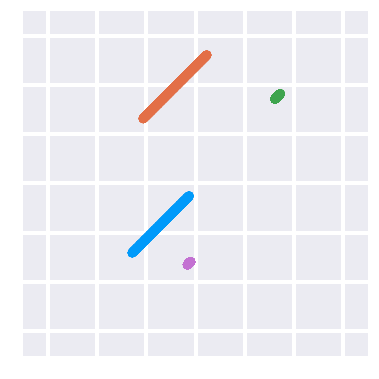}}
    \subfloat[\label{fig:llengths:hand}]{
        \includegraphics[width=0.245\linewidth]{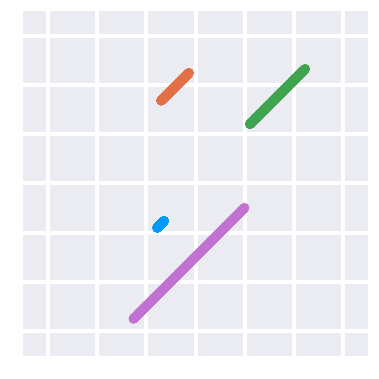}}

    \caption{Line lengths for different definitions of $l()$: a) the
    default, using the folded normal distribution; b) using the Poisson
    distribution, with $\lambda=l$; c) using the uniform distribution in the
    interval $\left[0, 2l\right[$; and, d) hand-picked lengths, more
    specifically $\pmb{\ell}=(2, 8, 16, 32)$.
    Cluster centers, as well as parameters $l$ and $l_\sigma$, are the same as
    for the example shown in Fig.~\ref{fig:flow}.}
    \label{fig:llengths}
\end{figure}

The $l()$ function reference is an optional parameter, allowing users to
customize how line lengths are obtained. By default, $l()$ references the
\texttt{llengths()} function, which determines the $\ell_i$ length of each
cluster-supporting line $i$ according to:

\begin{equation}
    \ell_i\sim\left|\mathcal{N}(l,l_\sigma^2)\right|
\end{equation}

\noindent where $\left|\mathcal{N}(\mu,\sigma^2)\right|$ represents the folded
normal distribution with mean $\mu$ and variance $\sigma^2$.
Fig.~\ref{fig:llengths} shows cluster-supporting line lengths obtained with
different definitions of $l()$.

\subsubsection{Determine angles between $\mathbf{d}$ and cluster-supporting lines}
\label{sec:methods:alg:angles}

The angles between the average direction of cluster-supporting lines,
$\mathbf{d}$, and the cluster-supporting lines themselves, are given by the
function referenced in $\theta_\Delta()$ according to:

\begin{equation}
    \mathbf{\Theta_\Delta} = \theta_\Delta(c, \theta_\sigma)
\end{equation}

\noindent where $\mathbf{\Theta_\Delta}$ is a $c$-dimensional vector containing
the final angle differences between $\mathbf{d}$ and the cluster-supporting
lines, $c$ is the number of clusters, and $\theta_\sigma$ is the angle
dispersion.

The $\theta_\Delta()$ function reference is an optional parameter, allowing
users to customize its behavior. By default, $\theta_\Delta()$ is specified by
the \texttt{angle\_deltas()} function, which determines the $\theta_{\Delta i}$
angle difference between $\mathbf{d}$ and the $i$-th cluster-supporting line
according to:

\begin{equation}
    \theta_{\Delta i}\sim\mathcal{WN}_{-\pi/2}^{\pi/2}(0,\theta_\sigma^2)
\end{equation}

\noindent where $\mathcal{WN}_{-\pi/2}^{\pi/2}(\mu,\sigma^2)$ represents the
wrapped normal distribution with mean $\mu$, variance $\sigma^2$, and support in
the $\left[-\pi/2,\pi/2\right]$ interval; $\theta_\sigma$ is the angle
dispersion of the cluster-supporting lines, specified as a mandatory parameter
to the \texttt{clugen()} function (see Table~\ref{tab:paramsmand}).

\begin{figure}[]
    \centering

    \subfloat[\label{fig:angles:wnormal}]{
        \includegraphics[width=0.245\linewidth]{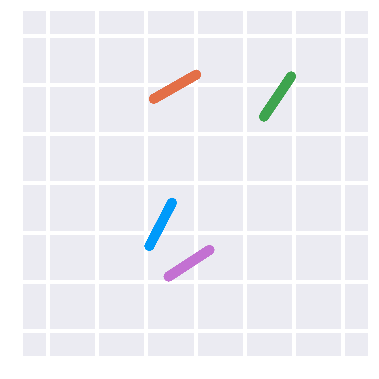}}
    \subfloat[\label{fig:angles:hand}]{
        \includegraphics[width=0.245\linewidth]{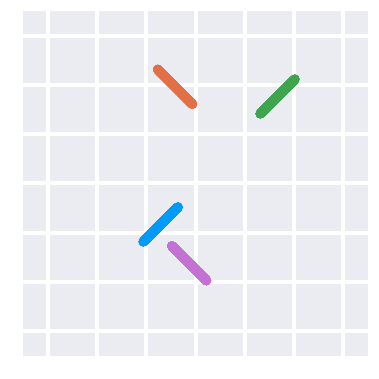}}

    \caption{Final directions of the cluster-supporting lines for two
    definitions of $\theta_\Delta()$: a) the default, where angle differences
    were obtained using the wrapped normal distribution; and, b) hand-picked
    angle differences, more specifically
    $\mathbf{\Theta_\Delta}=(0, \frac{\pi}{2}, 0, \frac{\pi}{2})$.
    Cluster centers, as well as the angle dispersion $\theta_\sigma$, are the same
    as for the example shown in Fig.~\ref{fig:flow}.}
    \label{fig:angles}
\end{figure}

Fig.~\ref{fig:angles} shows the final direction of the cluster-supporting lines
for two different definitions of $\theta_\Delta()$.

\subsubsection{Determine final directions of the cluster-supporting lines}
\label{sec:methods:alg:directions}

To obtain the $\hat{\mathbf{d}}_i$ final direction of each cluster-supporting
line $i$, the following algorithm is used:

\begin{enumerate}
    \item Find random vector $\mathbf{r}$ with each component taken from the
        uniform distribution between -0.5 and 0.5.
    \item Normalize $\mathbf{r}$:
        $\hat{\mathbf{r}}=\cfrac{\mathbf{r}}{\left\lVert\mathbf{r}\right\rVert}$.
    \item If $|\theta_{\Delta i}| > \pi/2$ or $n=1$, set
        $\hat{\mathbf{d}}_i=\hat{\mathbf{r}}$ and terminate the algorithm.
    \item If $\hat{\mathbf{r}}$ is parallel to $\hat{\mathbf{d}}$ go to 1.
    \item Determine vector $\mathbf{d}_\perp$ orthogonal to $\hat{\mathbf{d}}$
        using the first iteration of the Gram-Schmidt process:
        $\mathbf{d}_\perp=\hat{\mathbf{r}}-\cfrac{\hat{\mathbf{d}}\cdot\hat{\mathbf{r}}}{\hat{\mathbf{d}}\cdot\hat{\mathbf{d}}}\:\hat{\mathbf{d}}$.
    \item Normalize $\mathbf{d}_\perp$:
        $\hat{\mathbf{d}}_\perp=\cfrac{\mathbf{d}_\perp}{\left\lVert\mathbf{d}_\perp\right\rVert}$.
    \item Determine vector $\mathbf{d}_i$ at angle $\theta_{\Delta i}$ with
        $\hat{\mathbf{d}}$:
        $\mathbf{d}_i=\hat{\mathbf{d}}+\tan(\theta_{\Delta i})\hat{\mathbf{d}}_\perp$.
    \item Normalize $\mathbf{d}_i$:
        $\hat{\mathbf{d}}_i=\cfrac{\mathbf{d}_i}{\left\lVert\mathbf{d}_i\right\rVert}$.
\end{enumerate}

\subsubsection{Distance of point projections from the center of the
cluster-supporting line}
\label{sec:methods:alg:projs}

The distance of point projections from the center of their respective
cluster-supporting line $i$ is given by the function referenced in
$p_\text{proj}()$ according to:

\begin{equation}
    \mathbf{w}_i = p_\text{proj}(\ell_i, p_i)
    \label{eq:projs}
\end{equation}

\noindent where $\mathbf{w}_i$ is an $p_i \times 1$ vector containing the
distance of each point projection to the center of the line, while $\ell_i$ and
$p_i$ are the line length and number of points in cluster $i$, respectively.

The $p_\text{proj}()$ function reference is an optional parameter, allowing
users to customize how projections are placed on the cluster-supporting line.
Two concrete functions are provided out of the box by the software
implementations: \texttt{"norm"} and \texttt{"unif"}. The \texttt{clugen()}
function accepts these strings or a function reference compatible
with Eq.~\ref{eq:projs} as valid values, through its \texttt{proj\_dist\_fn}
parameter. The included functions work as follows:

\begin{description}
    \item[\texttt{"norm"}] (default) Each element of $\mathbf{w}_i$ is derived
        from $\mathcal{N}(0, (\frac{\ell_i}{6})^2)$, i.e., from the normal
        distribution, centered on the cluster-supporting line center ($\mu=0$)
        and with a standard deviation of $\sigma=\frac{\ell_i}{6}$, such that
        the length of the line segment encompasses $\approx$ 99.73\% of the
        generated projections. Consequently, some projections may be placed
        outside the line's endpoints.
    \item[\texttt{"unif"}] Each element of $\mathbf{w}_i$ is derived from
        $\mathcal{U}(-\frac{\ell_i}{2}, \frac{\ell_i}{2})$, i.e., from the
        continuous uniform distribution between
        $-\frac{\ell_i}{2}$ and $\frac{\ell_i}{2}$.
        Thus, projections will be uniformly dispersed along the
        cluster-supporting line.

\end{description}

\begin{figure}[]
    \centering

    \subfloat[\label{fig:pdist:norm}]{
        \includegraphics[width=0.245\linewidth]{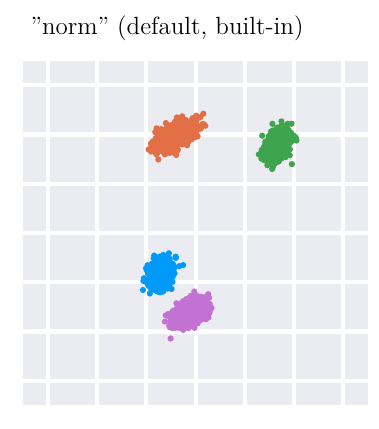}}
    \subfloat[\label{fig:pdist:unif}]{
        \includegraphics[width=0.245\linewidth]{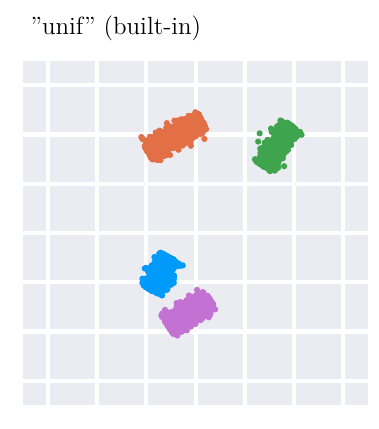}}
    \subfloat[\label{fig:pdist:laplace}]{
        \includegraphics[width=0.245\linewidth]{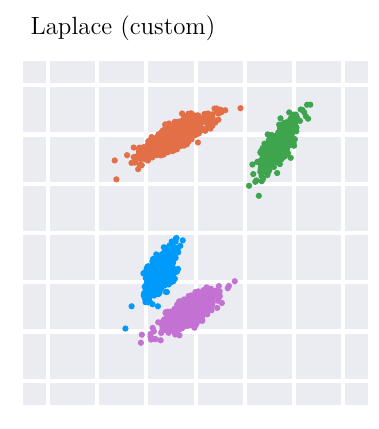}}
    \subfloat[\label{fig:pdist:rayleigh}]{
        \includegraphics[width=0.245\linewidth]{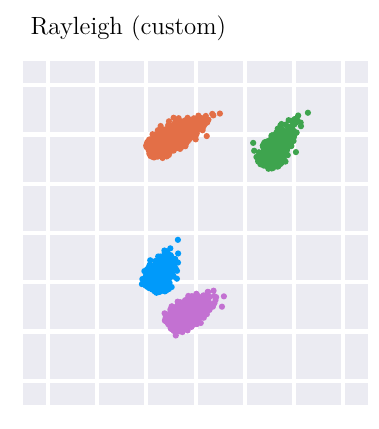}}

    \caption{Clusters generated for various definitions of $p_\text{proj}()$:
    a) the default, where line center distances are drawn for the normal
    distribution, specified using the in-built \texttt{"norm"} option;
    b) distances to the line center are derived from the uniform distribution,
    via the in-built \texttt{"unif"} option;
    c) line center distances are obtained from a custom user function
    implementing the Laplace distribution ($\mu=0, b=\frac{\ell_i}{6}$);
    and, d) custom user function returning center distances drawn from the
    Rayleigh distribution ($\sigma=\frac{\ell_i}{3}$) leaning on one of the
    corresponding line edges.
    All parameters are set as in Fig.~\ref{fig:flow}, except for
    $p_\text{proj}()$ in the case of (b)--(d), and $p$, which is set to 5000.}
    \label{fig:pdist}
\end{figure}

The impact of various definitions of $p_\text{proj}()$ is demonstrated in
Fig.~\ref{fig:pdist}. Fig.~\ref{fig:pdist:norm} and Fig.~\ref{fig:pdist:unif}
display clusters generated with the \texttt{"norm"} and \texttt{"unif"} options,
respectively. In turn, Fig.~\ref{fig:pdist:laplace} and
Fig.~\ref{fig:pdist:rayleigh} highlight custom user functions implementing the
Laplace and Rayleigh distributions, respectively.

\subsubsection{Coordinates of point projections on the cluster-supporting line}
\label{sec:methods:alg:projcoords}

This is a deterministic step performed by the \texttt{points\_on\_line()}
function using the vector formulation of the line equation, as follows:

\begin{equation}
    \mathbf{P}_i^\text{proj}=\mathbf{1}\,\mathbf{c}_i^T + \mathbf{w}_i\hat{\mathbf{d}}_i^T
\end{equation}

\noindent where $\mathbf{P}_i^\text{proj}$ is the $p_i \times n$ matrix of point
projection coordinates on the line, $\mathbf{1}$ is a $p_i \times 1$ vector
with all entries equal to 1, $\mathbf{c}_i$ are the coordinates of the line
center ($n \times 1$ vector), $\mathbf{w}_i$ is the distance of each point
projection to the center of the line ($p_i \times 1$ vector obtained in the
previous step), and $\hat{\mathbf{d}}_i$ is the normalized direction of the
cluster-supporting line for cluster $i$.

\subsubsection{Points from their projections on the cluster-supporting line}
\label{sec:methods:alg:points}

The final points in the $i$-th cluster, obtained from their projections on
the respective cluster-supporting line, are given by the function referenced in
$p_\text{final}()$ according to:

\begin{equation}
    \mathbf{P}_i^\text{final} = p_\text{final}(\mathbf{P}_i^\text{proj}, f_\sigma, \ell_i, \hat{\mathbf{d}}_i, \mathbf{c}_i)
    \label{eq:points}
\end{equation}

\noindent where $\mathbf{P}_i^\text{final}$ is a $p_i \times n$ matrix
containing the final coordinates of the generated points, $\mathbf{P}_i^\text{proj}$
is the $p_i \times n$ matrix of projection coordinates (determined in the
previous step), and $f_\sigma$ is the lateral dispersion parameter. In turn,
$\ell_i$, $\hat{\mathbf{d}}_i$, and $\mathbf{c}_i$ are the length, direction, and
center of the $i$-th cluster-supporting line, respectively.

\begin{figure}[]
    \centering

    \subfloat[\label{fig:points_detail:n_1}]{
        \includegraphics[width=0.245\linewidth]{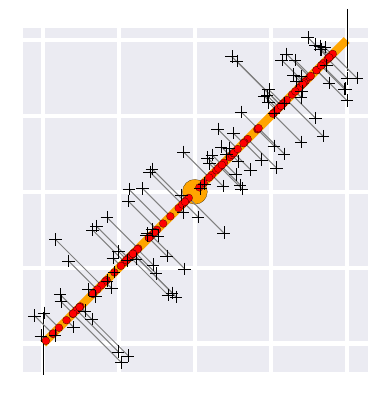}}
    \subfloat[\label{fig:points_detail:n}]{
        \includegraphics[width=0.245\linewidth]{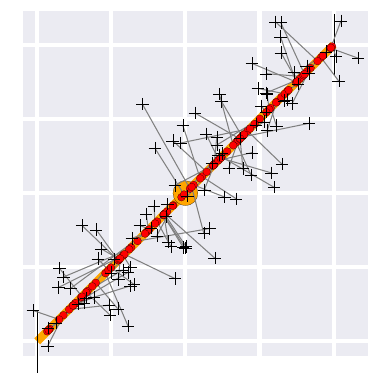}}

    \caption{Example of how the final cluster points are obtained in 2D when
    using the built-in implementations for $p_\text{final}()$:
    (a) \texttt{"n-1"}; and, (b) \texttt{"n"}.}
    \label{fig:points_detail}
\end{figure}

The $p_\text{final}()$ function reference is an optional parameter, allowing
users to customize how the final cluster points are generated from their
projections on the cluster-supporting line. Two concrete functions are provided
out of the box by the software implementations, specified by passing
\texttt{"n-1"} or \texttt{"n"} to \texttt{clugen()}'s \texttt{point\_dist\_fn}
parameter. Alternatively, the user can provide a custom function reference to
this parameter. The included functions work as follows:

\begin{description}
    \item[\texttt{"n-1"}] (default) Points are placed on a hyperplane orthogonal
        to the cluster-supporting line and intersecting the point's projection.
        This is done by obtaining $p_i$ random unit vectors orthogonal to
        $\hat{\mathbf{d}}_i$, and determining their magnitude using the normal
        distribution ($\mu=0$, $\sigma=f_\sigma$). These vectors are then added
        to the respective projections on the cluster-supporting line, yielding
        the final cluster points. Fig.~\ref{fig:points_detail:n_1} summarizes
        the process in 2D, where the orthogonal hyperplanes are simply lines
        perpendicular to the cluster-supporting line.
    \item[\texttt{"n"}] Points are placed around their respective projections.
        This is done by obtaining $p_i$ random unit vectors and determining
        their magnitude using the normal distribution ($\mu=0$,
        $\sigma=f_\sigma$). These vectors are then added to the respective
        projections on the cluster-supporting line, yielding the final cluster
        points, as shown in Fig.~\ref{fig:points_detail:n} for the 2D case.
\end{description}

\begin{figure}[]
    \centering

    \subfloat[\label{fig:points_norm:norm_n_1}]{
        \includegraphics[width=0.245\linewidth]{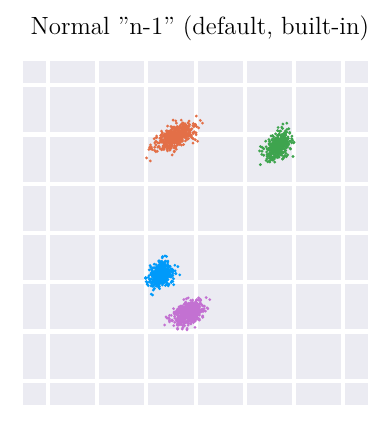}}
    \subfloat[\label{fig:points_norm:rician_n_1}]{
        \includegraphics[width=0.245\linewidth]{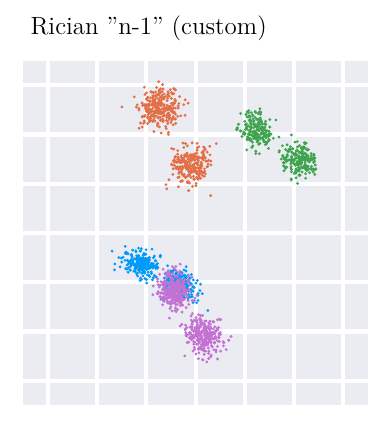}}
    \subfloat[\label{fig:points_norm:exp_n_1}]{
        \includegraphics[width=0.245\linewidth]{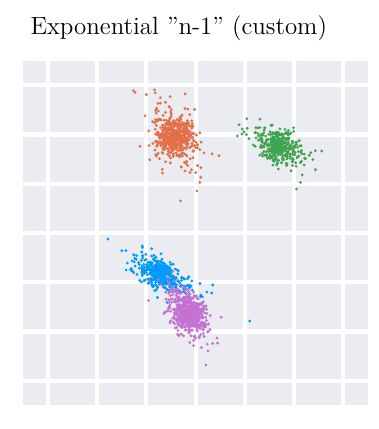}}
    \subfloat[\label{fig:points_norm:poisson_n_1}]{
        \includegraphics[width=0.245\linewidth]{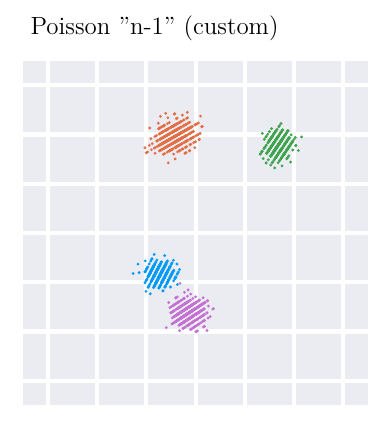}}

    \subfloat[\label{fig:points_norm:norm_n}]{
        \includegraphics[width=0.245\linewidth]{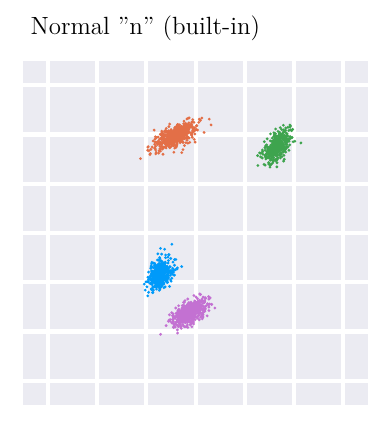}}
    \subfloat[\label{fig:points_norm:rician_n}]{
        \includegraphics[width=0.245\linewidth]{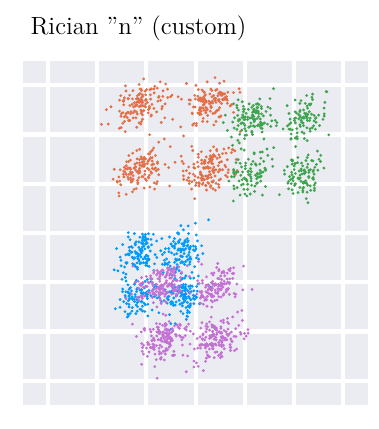}}
    \subfloat[\label{fig:points_norm:exp_n}]{
        \includegraphics[width=0.245\linewidth]{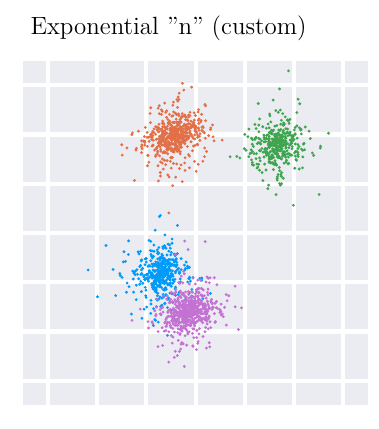}}
    \subfloat[\label{fig:points_norm:poisson_n}]{
        \includegraphics[width=0.245\linewidth]{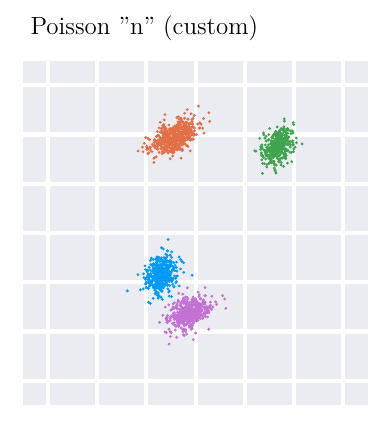}}

    \caption{Clusters generated for various definitions of $p_\text{final}()$.
    (a) \texttt{"n-1"}, which is the built-in default definition, corresponding
        to the normal distribution $(\mu=0,\sigma=f_\sigma)$ with an ``n-1''
        strategy;
    (b) Rician distribution $(\nu=\ell_i/2,\sigma=2f_\sigma)$ with an ``n-1''
        strategy (custom user function);
    (c) exponential distribution $(\lambda=f_\sigma/2)$ with an ``n-1''
        strategy (custom user function);
    (d) Poisson distribution $(\lambda=f_\sigma)$ with an ``n-1'' strategy
        (custom user function);
    (e) \texttt{"n"}, which is a built-in definition, corresponding to the
        normal distribution $(\mu=0,\sigma=f_\sigma)$ with an ``n'' strategy;
    (f) Rician distribution $(\nu=\ell_i/2,\sigma=2f_\sigma)$ with an ``n''
        strategy (custom user function);
    (g) exponential distribution $(\lambda=f_\sigma/2)$ with an ``n''
        strategy (custom user function);
    (h) Poisson distribution $(\lambda=f_\sigma)$ with a ``n'' strategy
    (custom user function).
    For (f)--(h), each final point coordinate is randomly multiplied by
    -1 or 1 so that points are placed around respective projections.
    The remaining parameters for all subfigures are defined as in
    Fig.\ref{fig:flow}, except for $p$, which is set to 5000.}
    \label{fig:points_norm}
\end{figure}

In general, points can be placed using an \textit{``n-1''} or \textit{``n''}
strategy using any distribution. Fig.~\ref{fig:points_norm} displays several
examples for various implementations of $p_\text{final}()$, either based on
\textit{``n-1''} or \textit{``n''} strategy, using different distributions.
Figures in the top row
(\ref{fig:points_norm:norm_n_1}--\ref{fig:points_norm:poisson_n_1}) display the
use of an ``n-1'' strategy, while figures in the bottom row
(\ref{fig:points_norm:norm_n}--\ref{fig:points_norm:poisson_n}) highlight
clusters obtained with an ``n'' strategy (the leftmost figure in each row,
Fig.~\ref{fig:points_norm:norm_n_1} and Fig.~\ref{fig:points_norm:norm_n},
correspond respectively to the build-in \texttt{"n-1"} and \texttt{"n"}
options). Figures in each column use a different distribution, namely
normal (\ref{fig:points_norm:norm_n_1} and \ref{fig:points_norm:norm_n}),
Rician (\ref{fig:points_norm:rician_n_1} and \ref{fig:points_norm:rician_n}),
exponential (\ref{fig:points_norm:exp_n_1} and \ref{fig:points_norm:exp_n}), and
Poisson (\ref{fig:points_norm:poisson_n_1} and \ref{fig:points_norm:poisson_n}).
The relation between the parameters of \texttt{clugen()} and the parameters used
for each distribution is specified in the figure's caption.

Several insights can be drawn by analyzing Fig.~\ref{fig:points_norm}. First,
note that, considering that Rician, exponential, and Poisson distributions
only generate non-negative values, the ``n-1'' strategy nonetheless places
points on both sides of the cluster-supporting lines. This occurs since the
generated random unit vectors perpendicular to $\hat{\mathbf{d}}_i$ can point in
opposite directions (in 3D and higher dimensions these vectors can indeed point
in infinite directions due to being placed in a hyperplane orthogonal to
$\hat{\mathbf{d}}_i$). Therefore, cluster-supporting lines remain as the
approximate centers of the respective clusters. Applying a non-negative
distribution directly with an ``n'' strategy would not have the same effect,
i.e., clusters would be placed somewhere in relation to the respective
cluster-supporting line, but not around it. Consequently, for
Figs.~\ref{fig:points_norm:rician_n}--\ref{fig:points_norm:poisson_n}, values
drawn from the corresponding distributions are randomly multiplied by either 1
or -1, which results in the cluster-supporting line continuing to be the center
of the respective cluster; in the case of the Rician distribution, which
generates average values above zero and is applied independently in each of the
two dimensions, this leads to four sub-clusters per cluster-supporting line,
one per quadrant, as shown in Fig.~\ref{fig:points_norm:rician_n}. This would
also happen with the Poisson distribution in Fig.~\ref{fig:points_norm:poisson_n}
if we increased $f_\sigma$ (thereby increasing the distribution's rate parameter,
$\lambda$, since $\lambda=f_\sigma$ in this example), moving the
average generated values farther away from zero.
Another interesting observation concerns the fact that the Poisson distribution
is discrete. This is clearly visible in Fig.~\ref{fig:points_norm:poisson_n_1},
in which points are placed at fixed distances from the cluster-supporting line.

\begin{figure}[]
    \centering

    \subfloat[\label{fig:points_unif:norm_n_1}]{
        \includegraphics[width=0.245\linewidth]{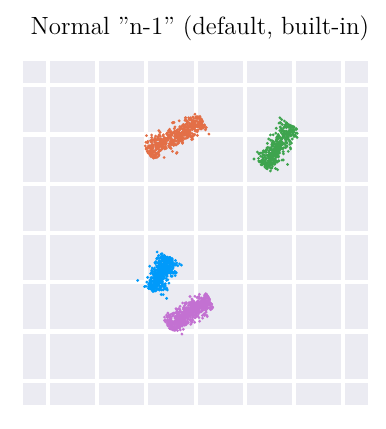}}
    \subfloat[\label{fig:points_unif:rician_n_1}]{
        \includegraphics[width=0.245\linewidth]{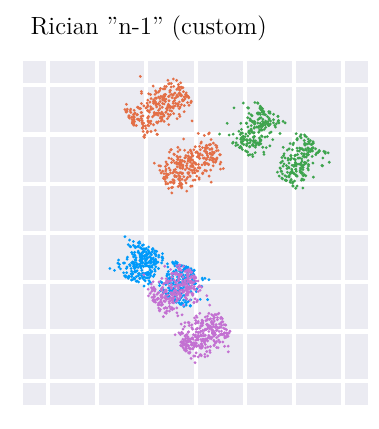}}
    \subfloat[\label{fig:points_unif:exp_n_1}]{
        \includegraphics[width=0.245\linewidth]{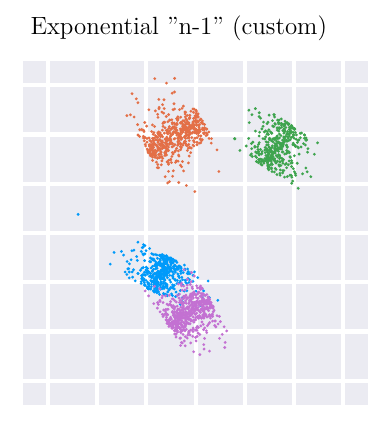}}
    \subfloat[\label{fig:points_unif:poisson_n_1}]{
        \includegraphics[width=0.245\linewidth]{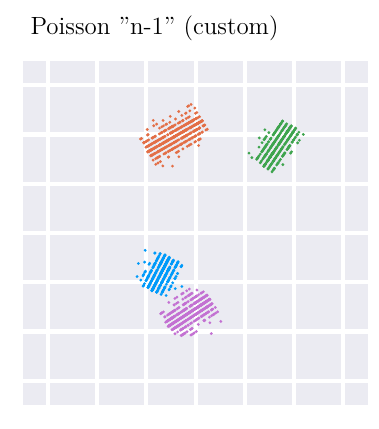}}

    \subfloat[\label{fig:points_unif:norm_n}]{
        \includegraphics[width=0.245\linewidth]{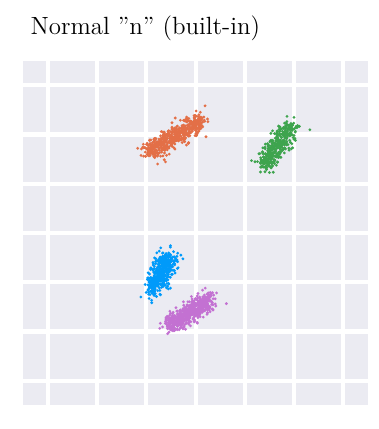}}
    \subfloat[\label{fig:points_unif:rician_n}]{
        \includegraphics[width=0.245\linewidth]{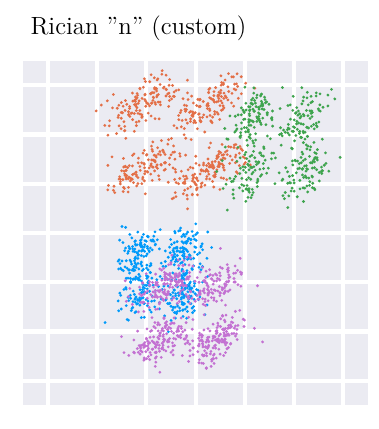}}
    \subfloat[\label{fig:points_unif:exp_n}]{
        \includegraphics[width=0.245\linewidth]{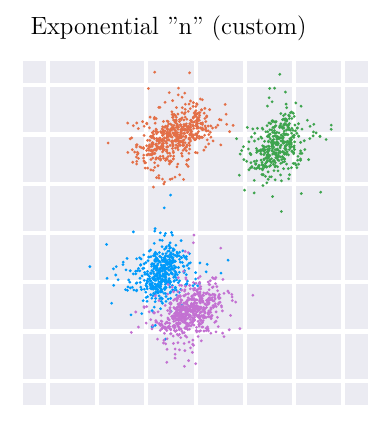}}
    \subfloat[\label{fig:points_unif:poisson_n}]{
        \includegraphics[width=0.245\linewidth]{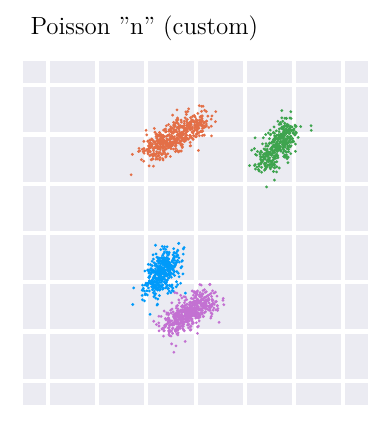}}

    \caption{Same as Fig.~\ref{fig:points_norm} but with \texttt{"unif"}
        projections.}
    \label{fig:points_unif}
\end{figure}

Overall, an ``n-1'' strategy allows the underlying point-placing distribution to
show more clearly, while an ``n'' strategy commonly generates more chaotic
clusters, in which the underlying point-placing distribution is not as obvious.
In any case, the decision of how to configure $p_\text{final}()$ is generally
not independent of $p_\text{proj}()$, and the user generally has to consider
the two parameters together to obtain the desired cluster shapes. For reference
purposes, Fig.~\ref{fig:points_unif} displays the same results shown in
Fig.~\ref{fig:points_norm}, but with $p_\text{proj}()$ set to \texttt{"unif"}.
Note that when $p_\text{proj}()$ is set to \texttt{"unif"} and $p_\text{final}()$
is set to \texttt{"n"} (Fig.~\ref{fig:points_unif:norm_n}), \textit{Clugen}
behaves similarly to Handl and Knowles ellipsoid generator
\cite{handl2005cluster}, minus the rejection of points located outside the
ellipsoid's boundary---which could, in any case, be implemented in a custom user
function.

\subsubsection{Output final clusters}
\label{sec:methods:alg:final}

The final step of the \textit{Clugen} algorithm consists of returning the
generated data points and the cluster groups to which they belong. However, the
\texttt{clugen()} function available in the four software implementations also
returns data from intermediate calculations, which may be useful in its own
right or for debugging and understanding the generation process. Generally
speaking, the \texttt{clugen()} function returns a structure or an object
containing the data described in Table~\ref{tab:output}.

\begin{table}[]
    \caption{Output of the \textit{Clugen} algorithm. The name column refers
        to field names in the object returned by the \texttt{clugen()}
        function in the several software implementations of the algorithm.}
    \label{tab:output}
    {\small
    \begin{tabular}{llp{10cm}}
    \toprule
    \multicolumn{2}{l}{Symbol / Name} & Description \\
    \midrule
    $\mathbf{P}^\text{final}$
        & \tabtt{points}
        & $p^\star \times n$ matrix with the coordinates of the generated
          points.\\
    $\mathbf{p_c}$
        & \tabtt{clusters}
        & $p^\star$-dim. integer vector associating points with clusters.\\
    $\mathbf{P}^\text{proj}$
        & \tabtt{projections }
        & $p^\star \times n$ matrix containing the coordinates of the generated
          projections on the cluster-supporting lines.\\
    $\mathbf{c_s}$
        & \tabtt{sizes}
        & $c$-dim. integer vector of cluster sizes such that
          $\sum{\mathbf{c_s}}=p^\star$.\\
    $\mathbf{C}$
        & \tabtt{centers}
        & $c \times n$ matrix with the final cluster centers.\\
    $\hat{\mathbf{D}}$
        & \tabtt{directions}
        & $c \times n$ matrix with the directions of cluster-supporting lines.\\
    $\mathbf{\Theta_\Delta}$
        & \tabtt{angles}
        & $c$-dim. vector with angle differences between $\mathbf{d}$ and
          the cluster-supporting lines.\\
    $\pmb{\ell}$
        & \tabtt{lengths}
        & $c$-dim. vector with lengths of cluster-supporting lines.\\

    \bottomrule
    \end{tabular}
    }
\end{table}

\subsection{Example Parameterizations in Higher Dimensions}
\label{sec:methods:examples}

The \textit{Clugen} algorithm was presented in detail in the previous
subsection, and 2D examples were used to exemplify several of its steps.
However, since the algorithm is capable of generating multidimensional data,
several higher-dimension parameterizations are additionally discussed in
this subsection.

\begin{figure}[]
    \centering
    {\tiny
    \begin{tabular}{cccccc}
        & & \multicolumn{4}{c}{{\normalsize $p_\text{final}()$}}\\
        & & & & & \\
        &
        & \texttt{"n-1"} (built-in, default)
        & \texttt{"n"} (built-in)
        & Pareto (custom)
        & Exponential (custom)
        \\
        \multirow[b]{13.3}{*}{\rotatebox[origin=c]{90}{\normalsize $p_\text{proj}()$}}
        &
        \rotatebox[origin=c]{90}{\texttt{"norm"} (built-in, default)}
        & \includegraphics[width=0.2\linewidth,align=c]{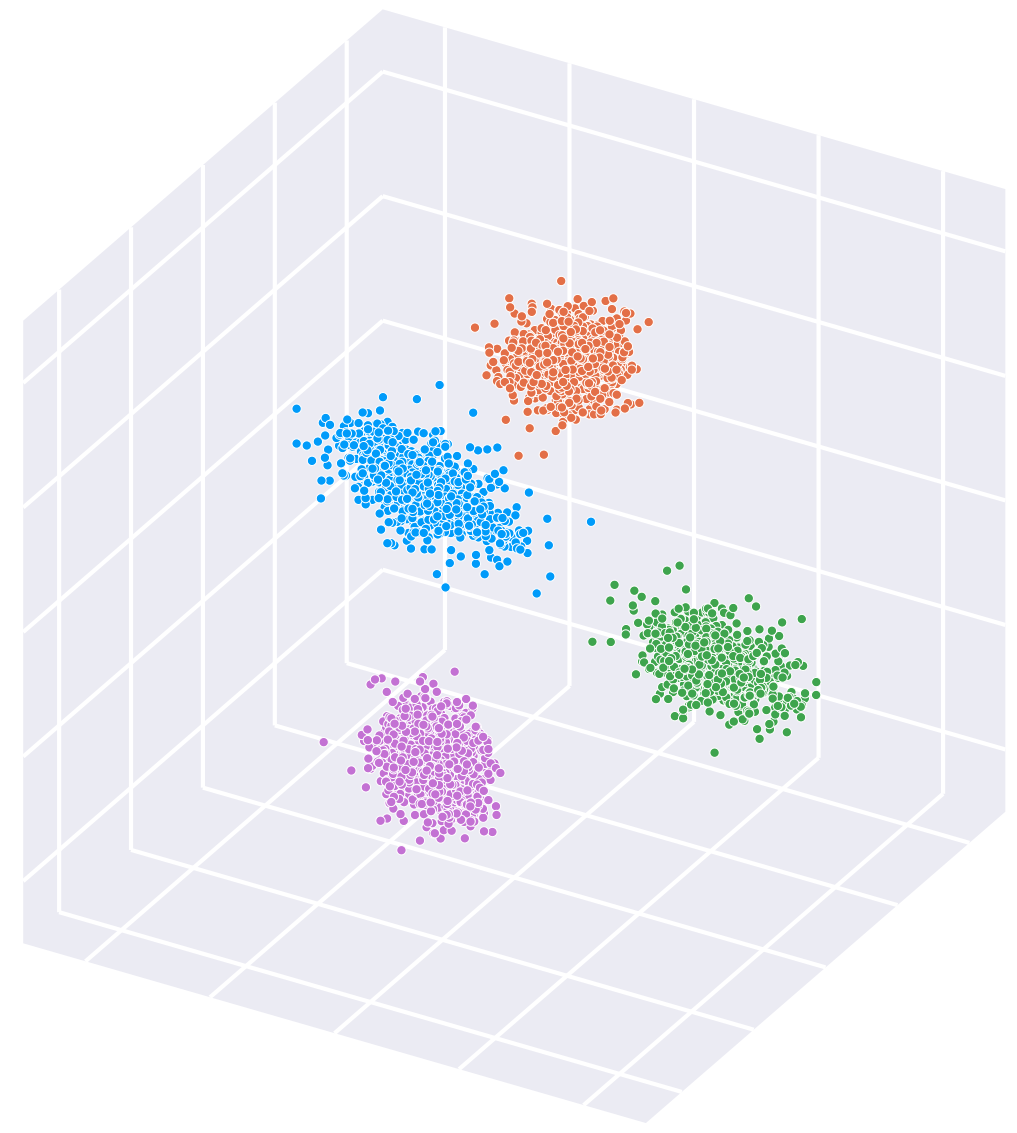}
        & \includegraphics[width=0.2\linewidth,align=c]{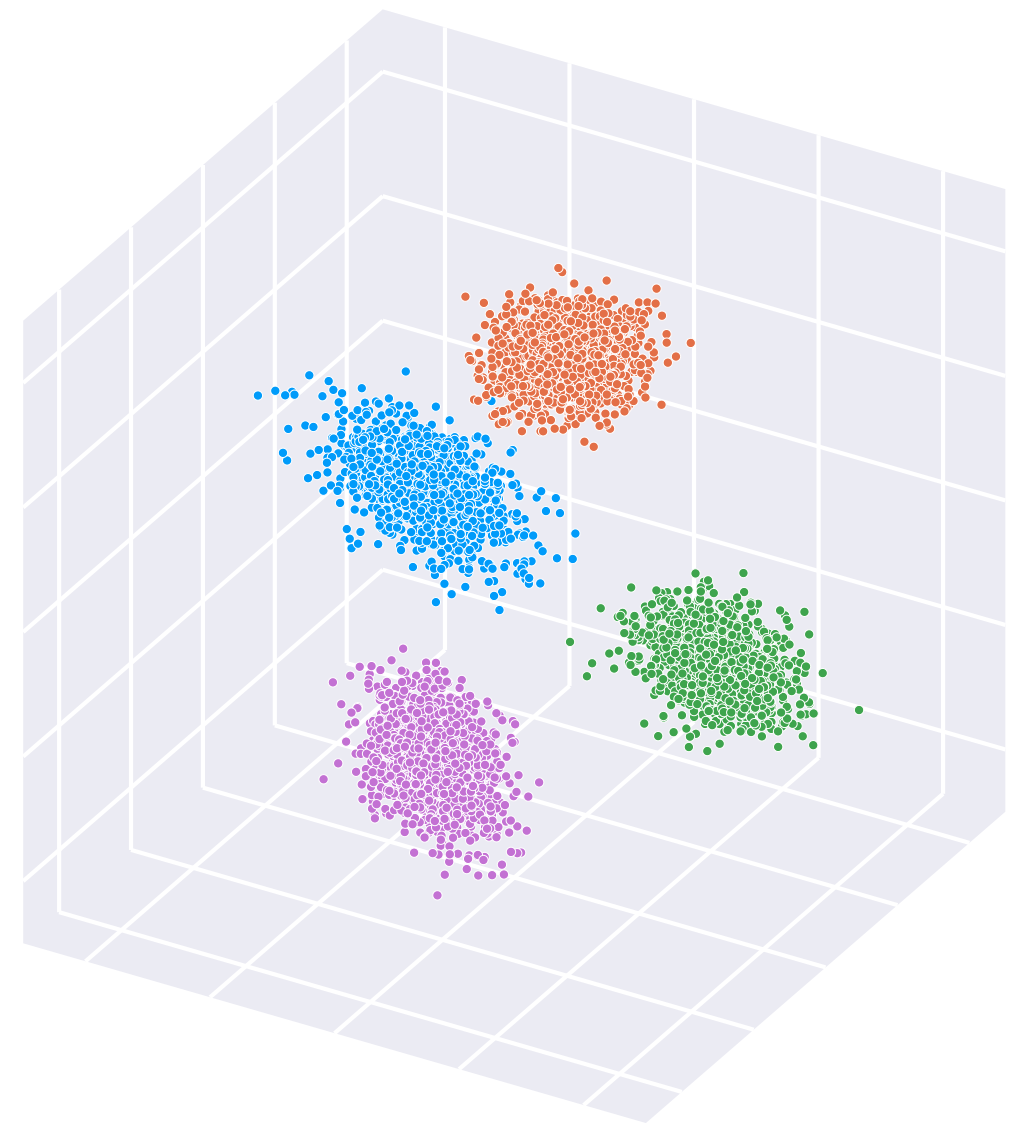}
        & \includegraphics[width=0.2\linewidth,align=c]{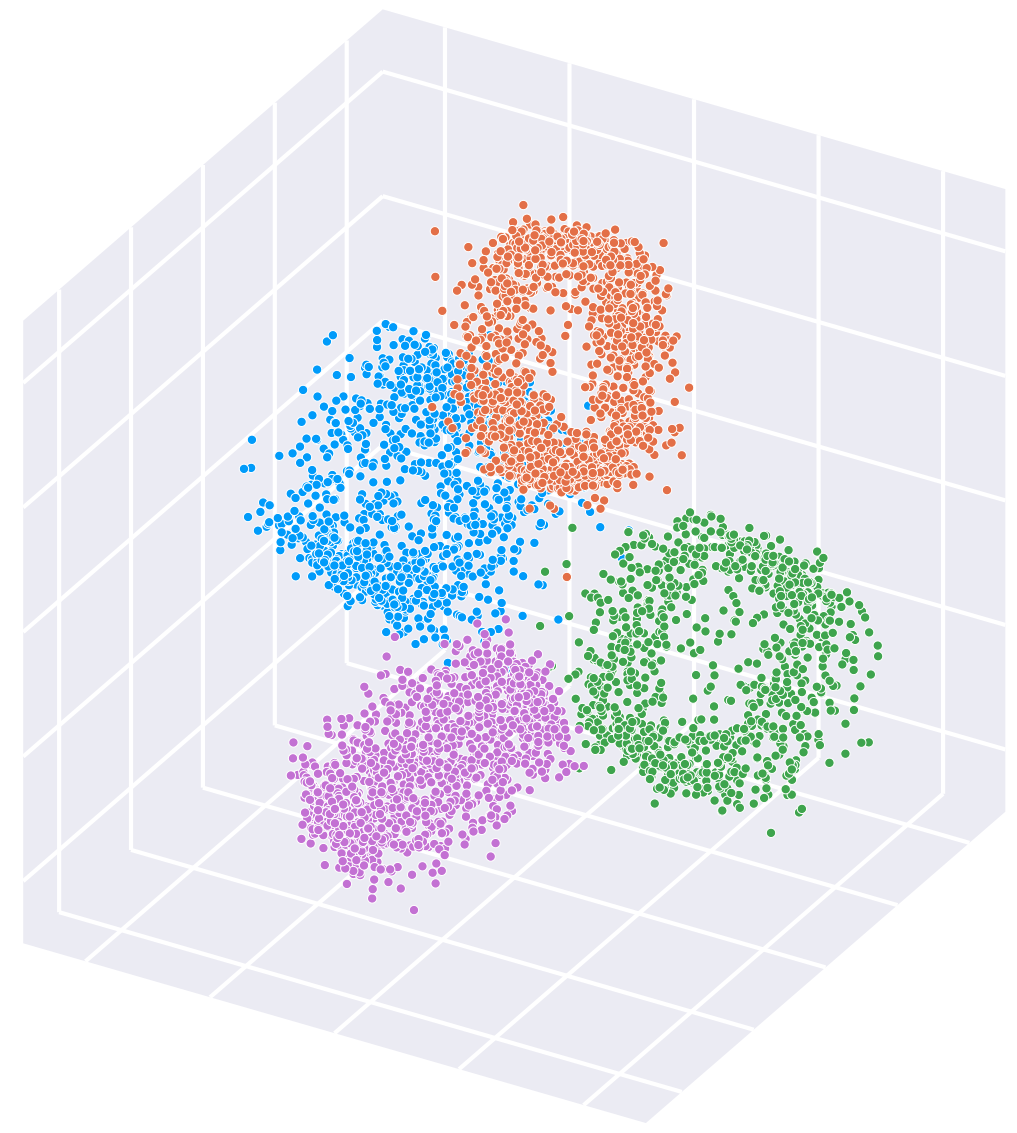}
        & \includegraphics[width=0.2\linewidth,align=c]{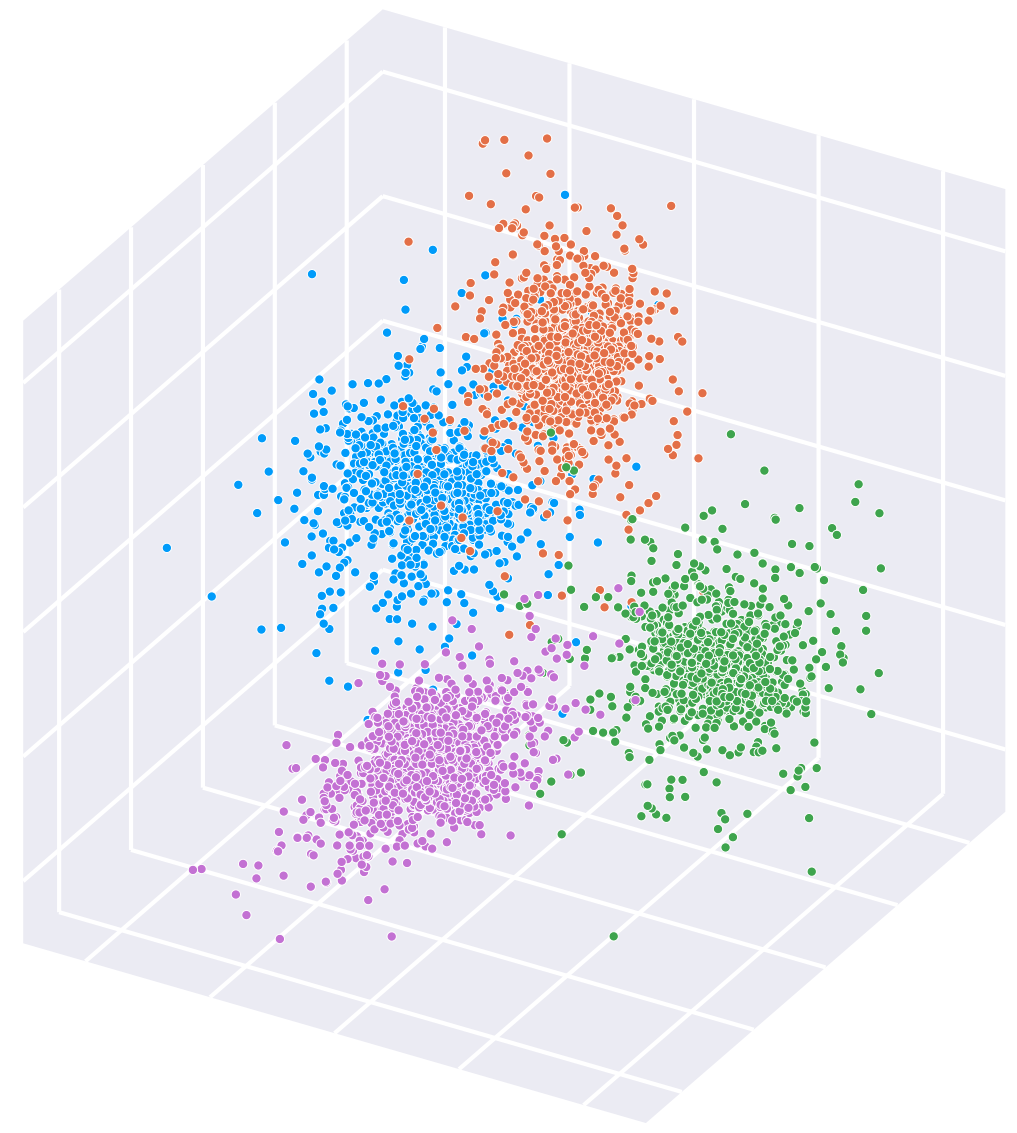}
        \\
        &
        \rotatebox[origin=c]{90}{\texttt{"unif"} (built-in)}
        & \includegraphics[width=0.2\linewidth,align=c]{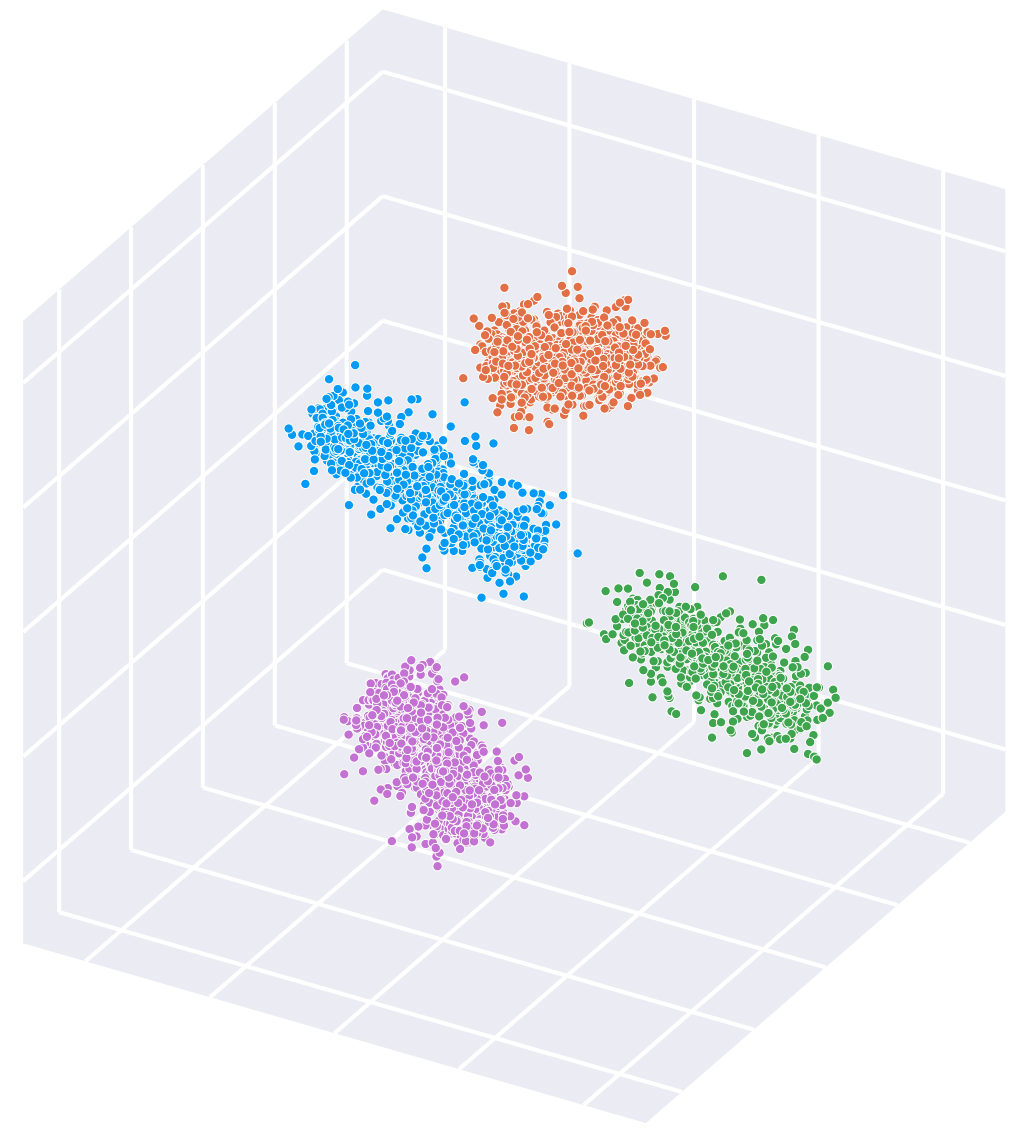}
        & \includegraphics[width=0.2\linewidth,align=c]{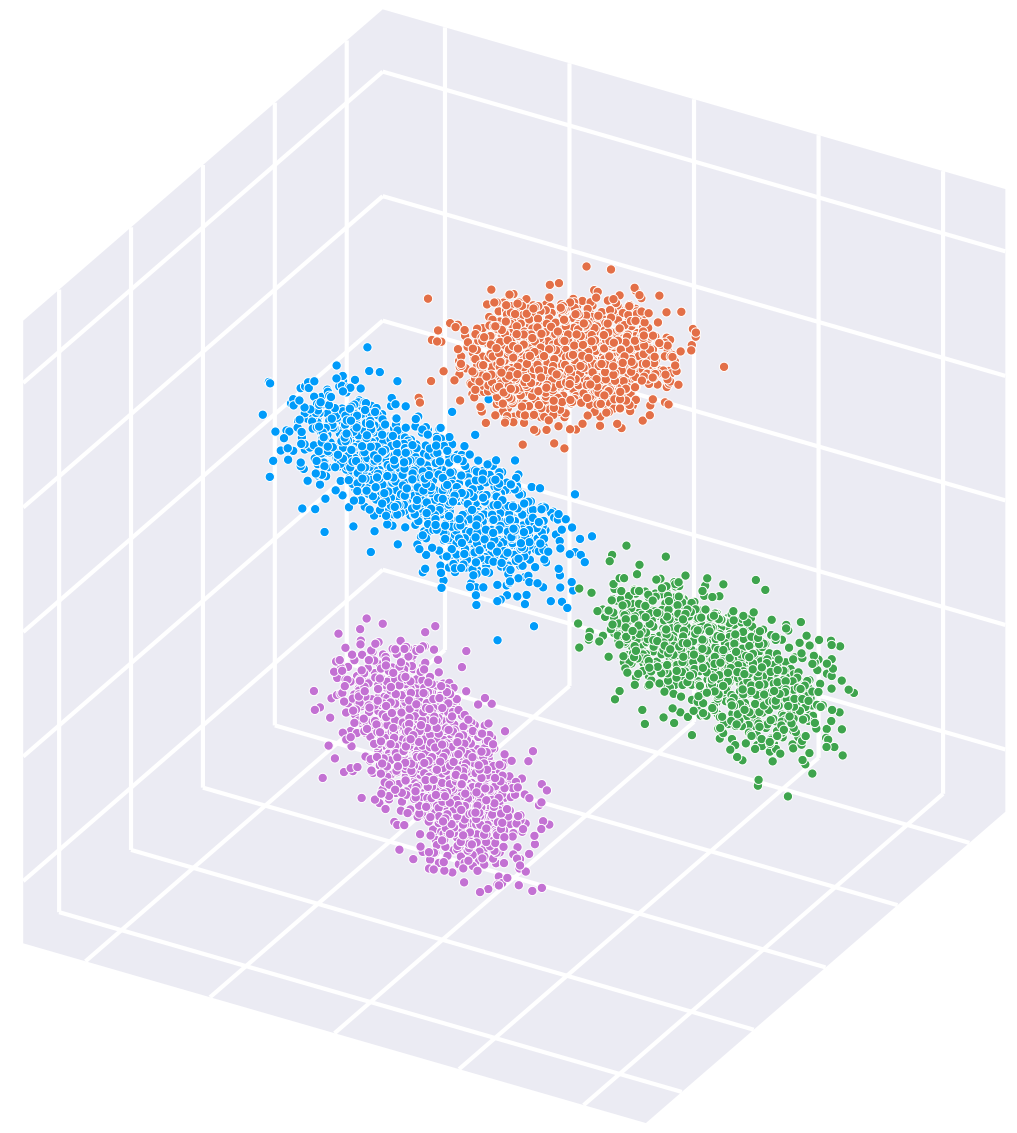}
        & \includegraphics[width=0.2\linewidth,align=c]{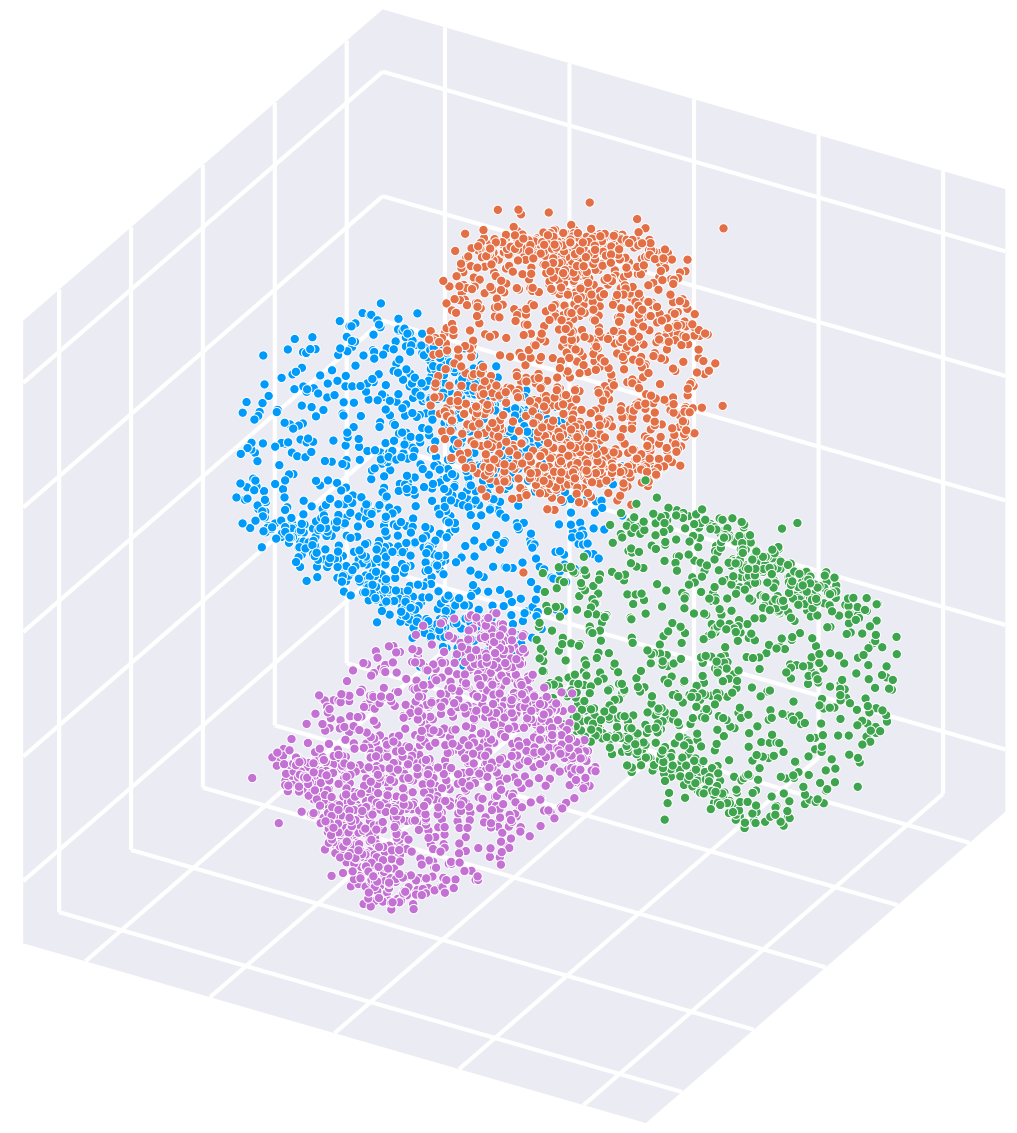}
        & \includegraphics[width=0.2\linewidth,align=c]{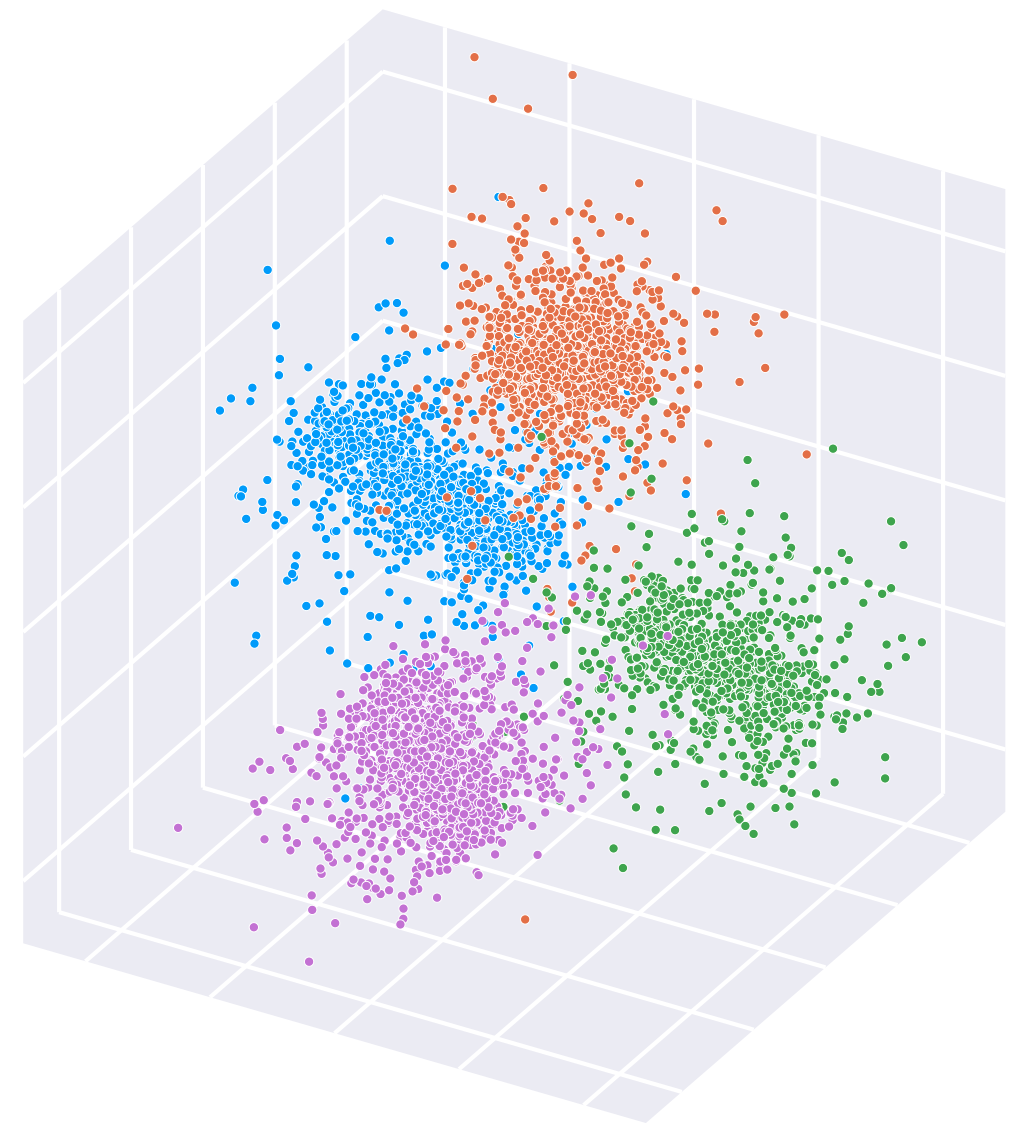}
        \\
        &
        \rotatebox[origin=c]{90}{Beta (custom)}
        & \includegraphics[width=0.2\linewidth,align=c]{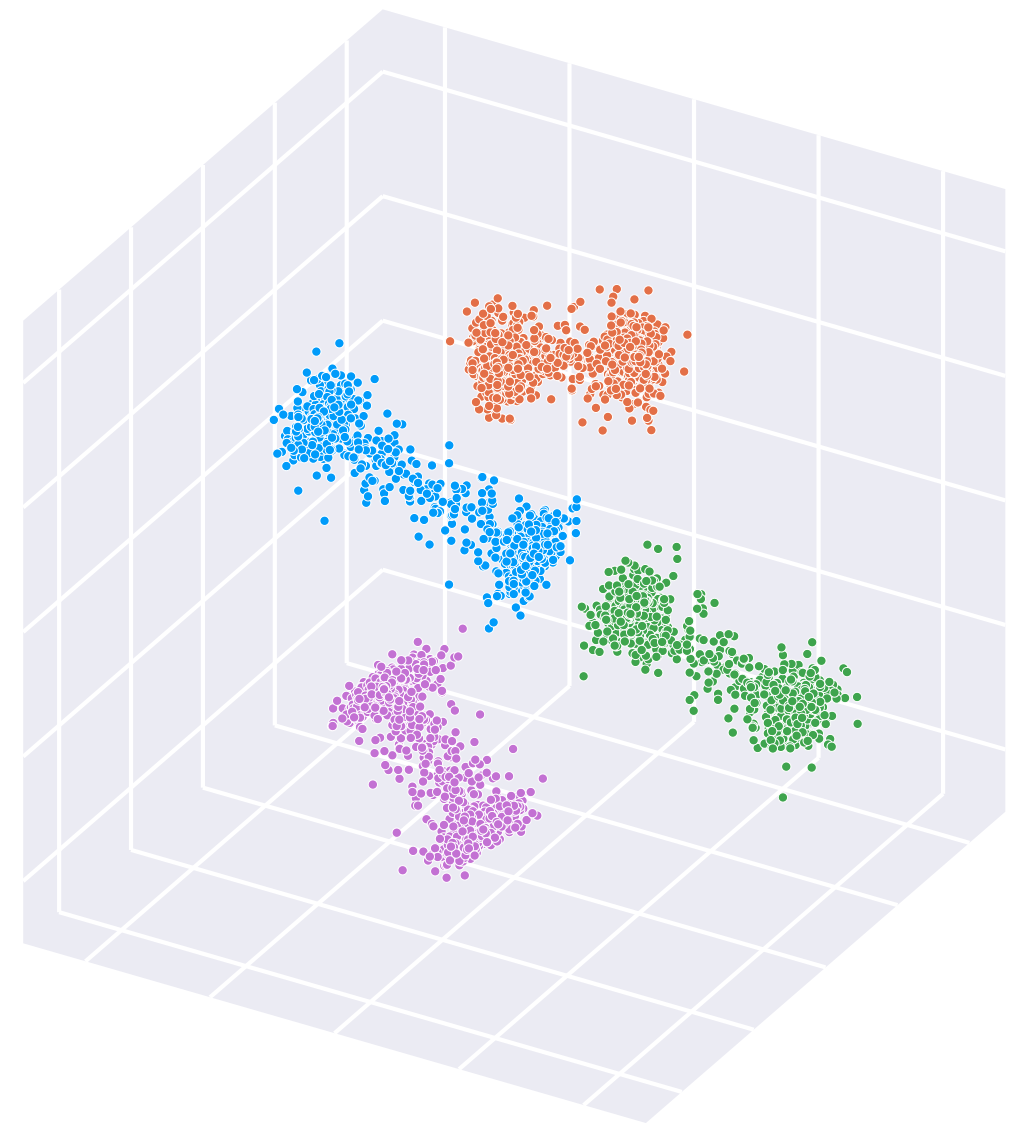}
        & \includegraphics[width=0.2\linewidth,align=c]{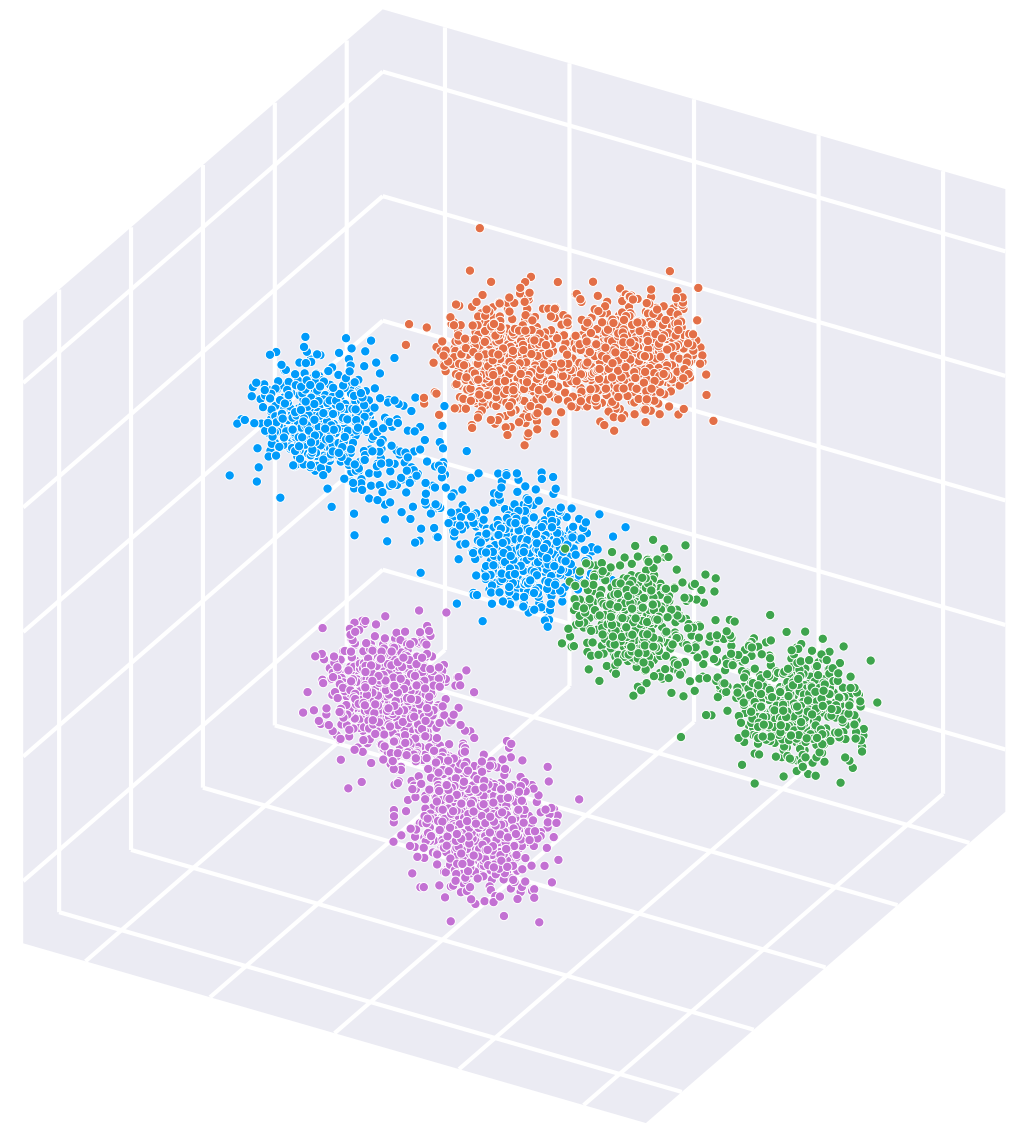}
        & \includegraphics[width=0.2\linewidth,align=c]{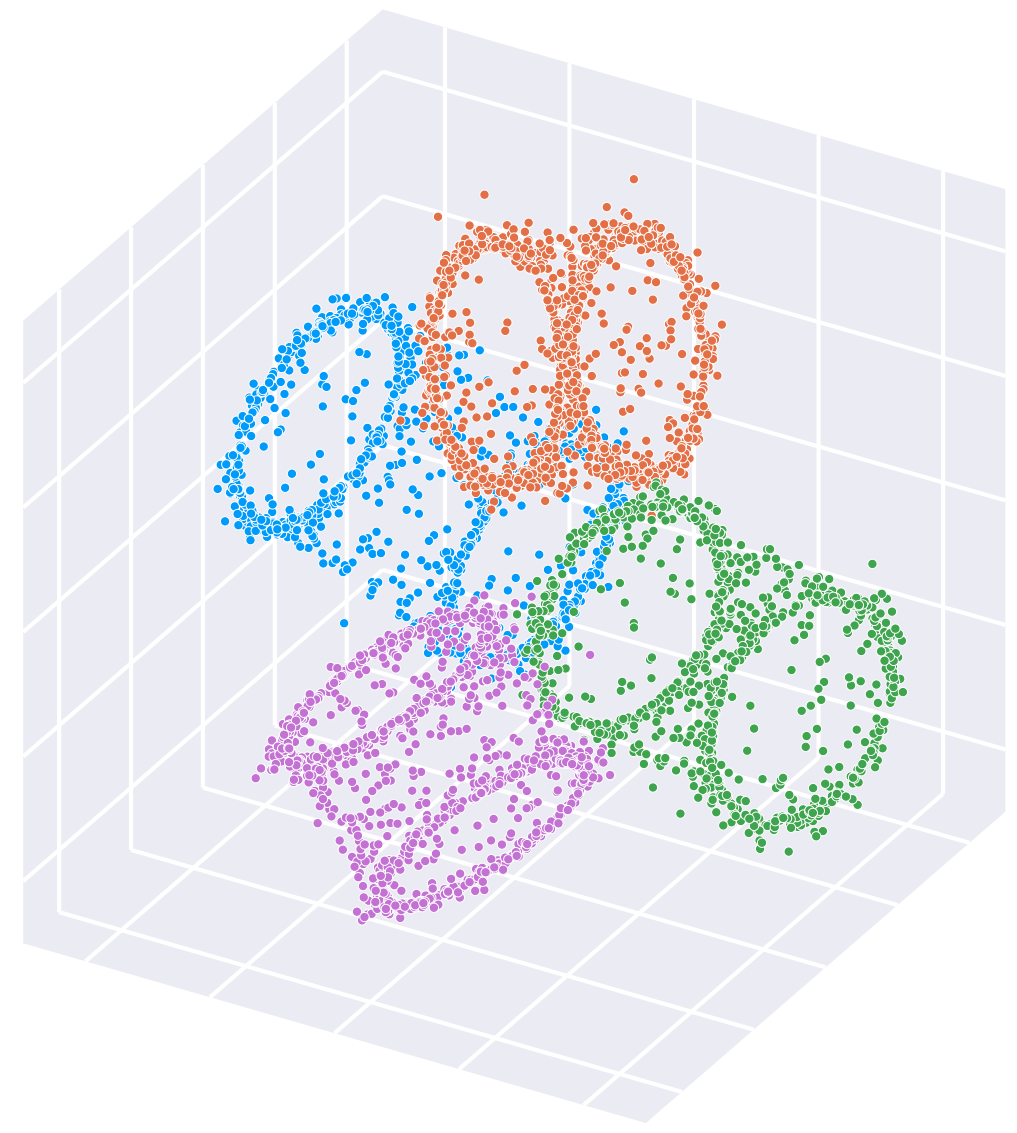}
        & \includegraphics[width=0.2\linewidth,align=c]{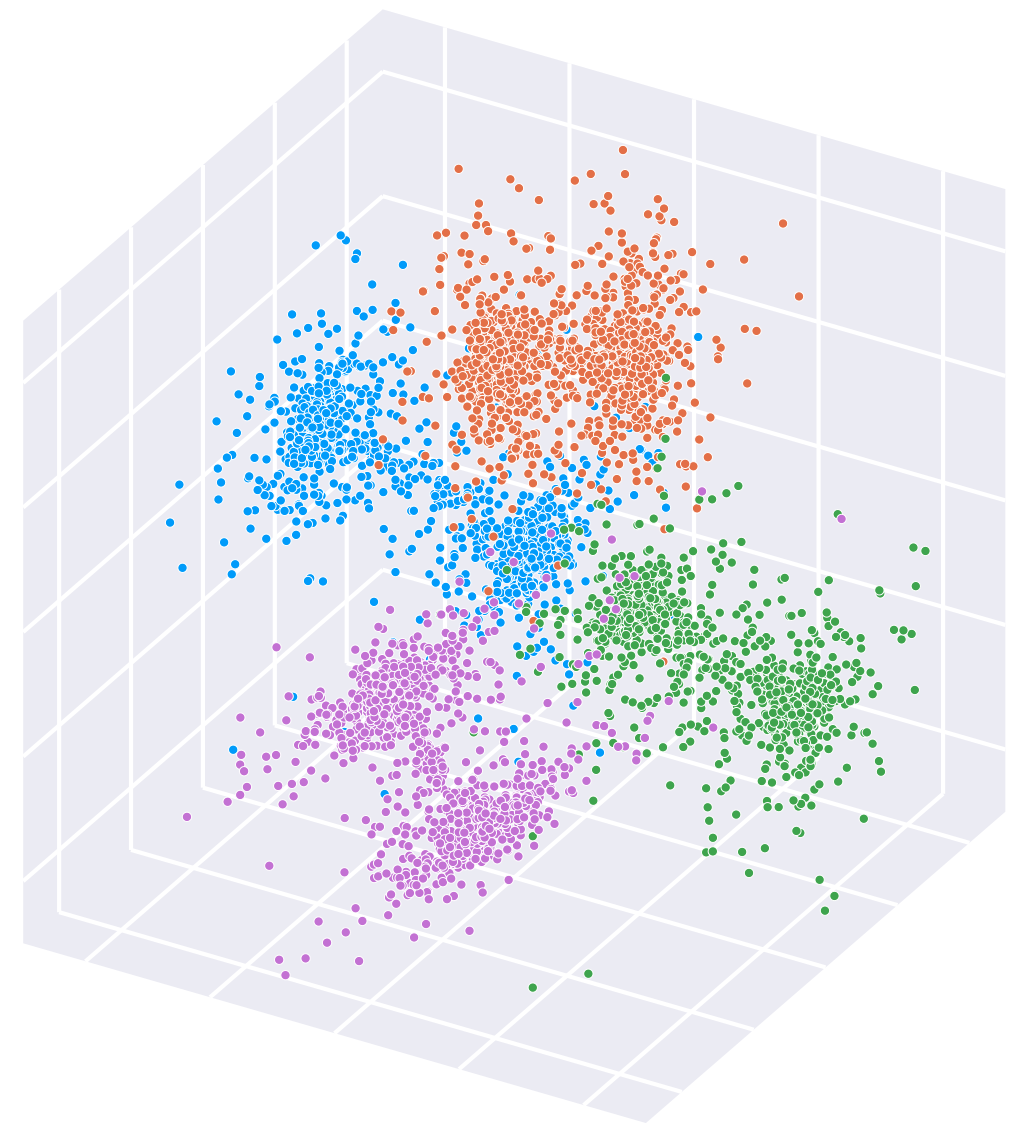}
    \end{tabular}}

    \caption{Several 3D examples of \textit{Clugen}-generated clusters with the
        following mandatory parameters: $c=4$, $p=5000$, $\mathbf{d}=(1,1,-1)$,
        $\theta_\sigma=\pi/4$, $\mathbf{s}=(6,6,6)$, $l=16$, $l_\sigma=3$, and
        $f_\sigma=2$.
        Rows correspond to different $p_\text{proj}()$ distributions, namely:
        \texttt{"norm"}, in which the point projections are normally
        distributed on the cluster-supporting lines; \texttt{"unif"}, where the
        point projections are uniformly distributed on the cluster-supporting
        lines; and, the Beta distribution $(\alpha=\beta=0.1)$, scaled to the
        length of the cluster-supporting lines. Columns correspond to various
        $p_\text{final}()$ distributions, specifically: \texttt{"n-1"}, in
        which final points are placed using the normal distribution
        $(\mu=0,\sigma=f_\sigma)$ with an ``n-1'' strategy; \texttt{"n"}, where
        final points are placed using the normal distribution
        $(\mu=0,\sigma=f_\sigma)$ with an ``n'' strategy; Pareto, where final
        points are placed using the Pareto distribution
        $(\alpha=l,x_\text{m}=4f_\sigma)$ using an ``n-1'' strategy; and,
        Exponential, for which the final points are placed using the
        Exponential distribution $(\lambda=f_\sigma/4)$ using an ``n-1''
        strategy.}
    \label{fig:ex3d}
\end{figure}

Fig.~\ref{fig:ex3d} shows several examples generated with \textit{Clugen}
in 3D. The mandatory parameters are the same in all images, while
$p_\text{proj}()$ and $p_\text{final}()$ vary across rows and columns,
respectively. Combinations of \texttt{"norm"} and \texttt{"unif"} projections
with \texttt{"n-1"}, \texttt{"n"} and Exponential final point placements simply
display a 3D perspective of the type of parameterizations already presented
in Fig.~\ref{fig:points_norm} and Fig.~\ref{fig:points_unif}. However, the use
of the Beta and Pareto distributions for $p_\text{proj}()$ and $p_\text{final}()$,
respectively, perfectly illustrates the immense possibilities offered by
\textit{Clugen}.

\begin{figure}[!t]
    \centering

    \includegraphics[width=1\textwidth]{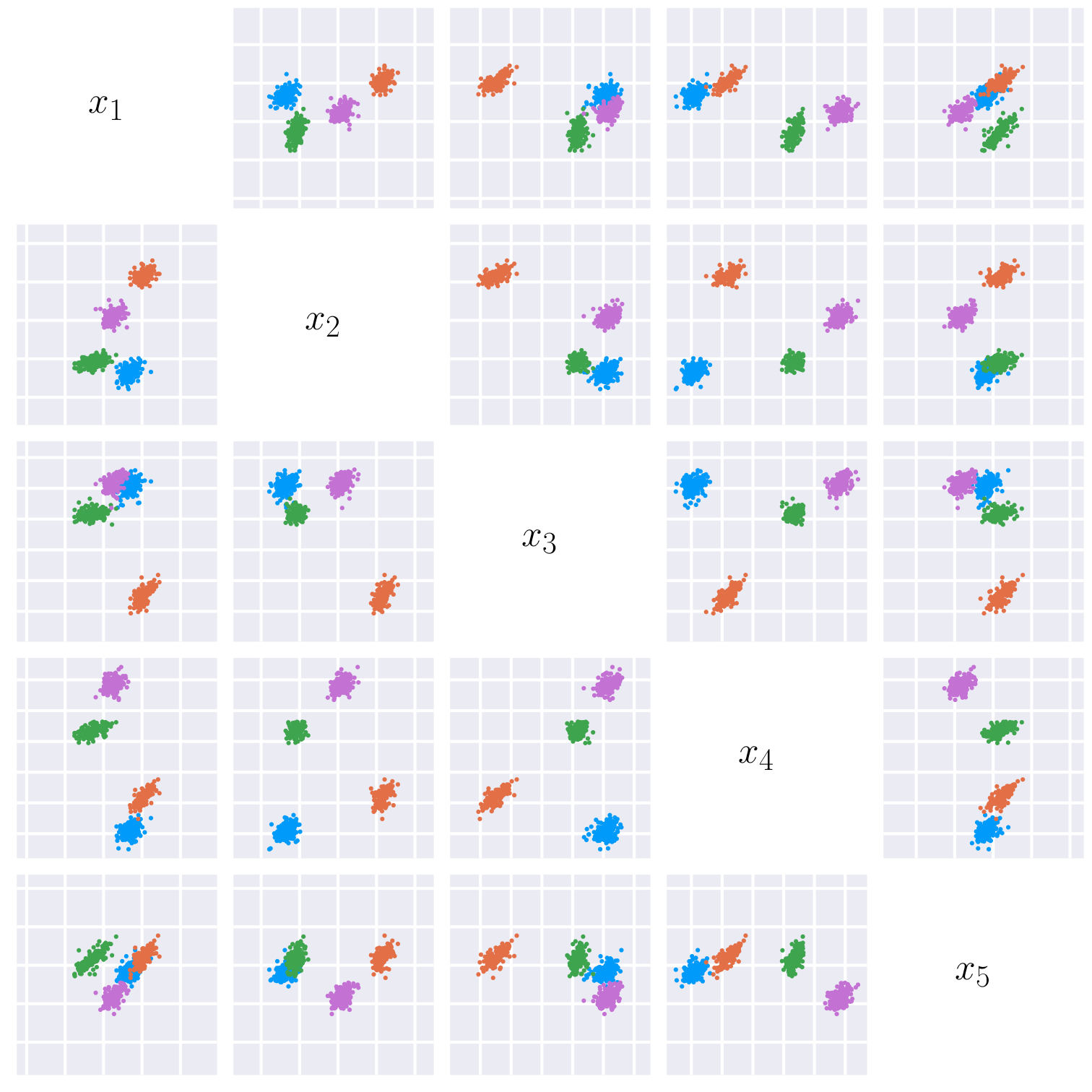}

    \caption{An example of \textit{Clugen}-generated clusters in 5D with the
    following mandatory parameters: $c=4$, $p=1000$, $\mathbf{d}=(1,1,1,1,1)$,
    $\theta_\sigma=\pi/10$, $\mathbf{s}=(30,30,30,30,30)$, $l=35$,
    $l_\sigma=10$, and $f_\sigma=4.5$. Optional parameters are set to their
    defaults.}
    \label{fig:ex5d}
\end{figure}

In turn, Fig.~\ref{fig:ex5d} shows an example in 5D, where each pair of
dimensions is plotted in the various subfigures. In this case, only the
mandatory parameters (indicated in the figure's caption) were specified, with
all optional parameters left to their defaults, highlighting the simplicity of
generating high-dimensional clusters in \textit{Clugen}.

\subsection{Software Implementations}
\label{sec:methods:impl}

Four implementations of the \textit{Clugen} algorithm are provided for the
Python, R, Julia, and MATLAB/Octave programming languages. The
implementations are available under the MIT license on GitHub at
\url{https://github.com/clugen}, and are comprehensively unit tested and
documented. A CI pipeline is used to regularly run the tests on the most recent
language ecosystem versions, as well as for publishing documentation and
packages. Table~\ref{tab:software} lists the implementations, providing their
names, minimum language requirements, and dependencies.

\begin{table}[]
\centering
    \caption{Software implementations.}
    \label{tab:software}
    {\small
    \begin{tabular}{lll}
    \toprule
    Name & Language & Runtime dependencies \\
    \midrule
    CluGen.jl& Julia ($\ge$ 1.1)& --- \\
    clugenr  & R ($\ge$ 3.6.0)  & mathjaxr ($\ge$ 1.2) \\
    MOCluGen & MATLAB ($\ge$ R2011a) & --- \\
                      & GNU Octave ($\ge$ 4.0.0) & --- \\
    pyclugen & Python ($\ge$ 3.8) & NumPy ($\ge$ 1.20) \\

    \bottomrule
    \end{tabular}
    }
\end{table}

The four implementations are functionally equivalent, leveraging the respective
platform's numerical computing framework. All of the platforms natively support
numerical computing, except for Python, although the third-party NumPy library
\cite{harris2020numpy} perfectly fits that role. The implementations follow a
procedural programming approach, using vectorization---i.e., operating on vectors
and matrices instead of the elements that comprise them---and avoiding copying
data in memory whenever possible. Further, the implementations have minimal
dependencies. The Python implementation depends on NumPy, and the R version
requires mathjaxr \cite{viechtbauer2022mathjaxr} for displaying equations in
the bundled user help.

However, there are some differences regarding the way implementations handle
reproducibility and the generation of pseudo-random numbers, essentially
reflecting the differences in the underlying platforms. For example, in
clugenr and MOCluGen---the R and MATLAB/Octave implementations of
\textit{Clugen}, respectively---reproducible executions of the different
stochastic functions require the user to seed a global pseudo-random number
generator (PRNG). These implementations allow passing an optional PRNG seed
to the main \texttt{clugen()} function to automatically perform this task,
simplifying the code. Conversely, for CluGen.jl and pyclugen---the Julia
and Python implementations, respectively---the user
needs to provide an initialized PRNG object to the different stochastic
functions for obtaining reproducibility. CluGen.jl offers the most
versatility since it is also possible to set the seed of Julia's global
PRNG, which the various functions will use by default if not provided
with a local PRNG object.

\begin{table}[]
    \caption{Functions offered by the software implementations of the
        \textit{Clugen} algorithm. The \texttt{clugen()} function
        performs the high-level steps of the algorithm. Functions in the
        Core group perform fundamental low-level operations. The Module group
        contains functions that implement the default behavior in several of the
        algorithm steps (shown in light blue in Fig.~\ref{fig:flow}). Finally,
        the Helper functions' code could have been
        incorporated in the previous functions; however, by existing as separate
        entities, these functions simplify testing and streamline the
        implementation of custom user functions for the Module group.}
    \label{tab:functions}
    {\small
    \begin{tabular}{llp{7cm}}
    \toprule
    Group & Function name & Description \\
    \midrule
    Main
        & \tabtt{clugen()} & Generates multidimensional clusters. \\
        & \tabtt{clumerge()} & Merges multidimensional data from arbitrary
            sources.\\
    \hdashline[.5pt/1pt]
    Core
        & \tabtt{points\_on\_line()} & Get position of points on a line
            given their distance to its center
            (Subsec.~\ref{sec:methods:alg:projcoords}).
        \\
        & \tabtt{rand\_ortho\_vector()} & Get a random unit vector orthogonal to
            the input vector. \\
        & \tabtt{rand\_unit\_vector()} & Get a random unit vector with the
            specified dimensions. \\
        & \tabtt{rand\_vector\_at\_angle()} & Get a random unit vector at a
            given angle with another vector. \\
    \hdashline[.5pt/1pt]
    Module
        & \tabtt{angle\_deltas()} & Obtain angles between $\mathbf{d}$
            and cluster-supporting lines (Subsec.~\ref{sec:methods:alg:angles}).
        \\
        & \tabtt{clucenters()} & Determine cluster centers
            (Subsec.~\ref{sec:methods:alg:centers}).
        \\
        & \tabtt{clupoints\_n\_1()} & Get final points when the
            \tabtt{point\_dist\_fn} parameter is set to \tabtt{"n-1"}
            (Subsec.~\ref{sec:methods:alg:points}).
        \\
        & \tabtt{clupoints\_n()} & Get final points when the
            \tabtt{point\_dist\_fn} parameter is set to \tabtt{"n"}
            (Subsec.~\ref{sec:methods:alg:points}).
        \\
        & \tabtt{clusizes()} & Determine cluster sizes, i.e., the number of
            points in each cluster (Subsec.~\ref{sec:methods:alg:clusizes}).\\
        & \tabtt{llengths()} & Determine length of cluster-supporting lines
            (Subsec.~\ref{sec:methods:alg:lengths}). \\
    \hdashline[.5pt/1pt]
    Helper
        & \tabtt{angle\_btw()} & Angle between two $n$-dimensional vectors.
        \\
        & \tabtt{clupoints\_n\_1\_template()} & Implements a generic ``n-1''
            strategy, supporting \tabtt{clupoints\_n\_1()} and custom user
            functions (Subsec.~\ref{sec:methods:alg:points}).
        \\
        & \tabtt{fix\_empty()} & Certifies that, given enough points, no
            clusters are left empty (Subsec.~\ref{sec:methods:alg:clusizes}).
        \\
        & \tabtt{fix\_num\_points()} & Certifies that array values add up to a
            specific total (Subsec.~\ref{sec:methods:alg:clusizes}).\\
    \bottomrule
    \end{tabular}
    }
\end{table}

The software implementations are organized into several functions, listed in
Table~\ref{tab:functions}. This separation of concerns fosters unit testing
and simplifies the creation of custom user functions, as it provides common
functionality in readily available building blocks. One example of this is
the \texttt{clupoints\_n\_1\_template()} function, which not only supports
\texttt{clupoints\_n\_1()} (which is the function invoked when the user
sets $p_\text{final}()$ to \texttt{"n-1"}), but also greatly streamlines
the creation of custom user functions following an ``n-1'' strategy.

In more specific cases, this separation of concerns also allows users to
entirely sidestep the main \texttt{clugen()} function and use the various
functions for composing their own cluster generation algorithms. Further,
some of the functions may be useful outside the context of synthetic data
generation, especially the ones that manipulate generic $n$-dimensional
vectors, such as \texttt{random\_vector\_at\_angle()} or
\texttt{angle\_btw()}.

In addition to implementing the \textit{Clugen} algorithm using modular and
interchangeable blocks, the four implementations also provide some
``quality of life'' features for users who need more detailed control over the
generated data. In particular, users can: i) directly specify the direction
of each cluster-supporting line (as an alternative to the average cluster
direction); ii) directly specify cluster sizes, cluster centers, line lengths,
and angle deltas (as an alternative to these being drawn from statistical
distributions); and, iii) merge datasets---created with \texttt{clugen()} or
otherwise---using the \texttt{clumerge()} function.

\section{Usage Example: Assessing Clustering Algorithms}
\label{sec:useexample}

As previously stated, a potential use of \textit{Clugen} is in generating many
cluster sets with similar characteristics to assess different clustering
algorithms. Further, given that one of the features of \textit{Clugen} is that
of generating arbitrarily elongated clusters, we present a usage example where
five different clustering algorithms are tested
with cluster sets containing increasingly elongated clusters. The example was
implemented in Julia, using CluGen.jl for cluster generation and the
Clustering.jl package \cite{clustering2012jl} for clustering and clustering
assessment. The example's reproducible code is available as supplementary
material \cite{fachada2022notebooks}.
The clustering algorithms tested, as well as the clustering quality assessment used,
are described in Subsection~\ref{sec:useexample:algs}. The experimental setup is
outlined in Subsection~\ref{sec:useexample:expsetup}. Results of this experiment
are presented and analyzed in Subsection~\ref{sec:useexample:results}, while
its limitations are discussed in Subsection~\ref{sec:useexample:limitations}.

\subsection{Clustering Algorithms and the $V$-measure Assessment Metric}
\label{sec:useexample:algs}

We tested five well-known clustering algorithms with data generated by
\textit{Clugen}, namely $k$-means, fuzzy $c$-means, $k$-medoids, agglomerative
hierarchical clustering (AHC), and density-based spatial clustering of
applications with noise (DBSCAN). While standard implementations of $k$-means
and fuzzy $c$-means assume input data in the Euclidean space, the remaining
algorithms---$k$-medoids, AHC, and DBSCAN---accept input data in the form of a
dissimilarity matrix, directly allowing the use of different distance metrics.

More specifically, $k$-means is a classical clustering algorithm where, given
$k$ cluster centroids, observations are iteratively assigned to their nearest
centroid \cite{hastie2009elements,lloyd1982least}. Centroids are then
recalculated such that they represent the mean of their new assignments. This
process continues until no more reassignments occur. The starting centroids can
be obtained in several ways, but for this study, we used $k$-means++
initialization, which reduces the running time of the algorithm while improving
the quality of the final solutions \cite{arthur2006k}.

The fuzzy $c$-means algorithm is relatively similar to $k$-means. The main
difference, however, is that all observations in the data contribute to
determining the cluster centroids according to their degree of belonging to the
respective clusters \cite{dunn1973fuzzy}. The $\mathsf{m}$ parameter, often
set to 2, controls the cluster membership fuzziness, with larger values
resulting in more clusters sharing their observations and vice versa (when
$\mathsf{m}\to\infty$, cluster membership is the same for all observations).
Conversely, fuzzy $c$-means becomes analogous to $k$-means when
$\mathsf{m}\to 1$ \cite{schwammle2010simple}.

In a similar fashion to $k$-means, the $k$-medoids algorithm works by finding
$k$ observations---the medoids, each representing a cluster---such that the
average dissimilarity between them and their assigned observations is minimized
\cite{hastie2009elements}. Unlike $k$-means, the medoids are actual
observations. Here, we experiment with a $k$-medoids version more in line with
$k$-means \cite{schubert2019faster} than its original version
\cite{kaufman1990partitioning}, allowing, for example, various medoid
initialization strategies, such as the aforementioned $k$-means++ (which we
use for this experiment).

In AHC \cite{hastie2009elements}, individual observations start in their own
cluster, and pairs of clusters are iteratively merged until there is only one
cluster containing all the data (or a desired number of clusters has been
reached). At each step, the cluster pair with the smallest intercluster
dissimilarity is selected for merging. This dissimilarity is determined by a
linkage criterion, which is directly related to the pairwise distances between
observations. Therefore, the linkage criterion ($\mathsf{l_{||}}$) and the
distance metric ($\mathsf{d_{||}}$) between pairs of observations are critical
parameters in AHC.

DBSCAN \cite{ester1996density} is a density-based clustering algorithm, grouping
points with sufficient neighbors in close proximity. The algorithm works by
identifying points in the neighborhood of every point---defined by a radius
$\mathsf{\epsilon}$---selecting points with $\mathsf{p_{min}}$ or more
neighbors. Points far from existing clusters and/or without enough close
neighbors are labeled as noise. Besides $\mathsf{\epsilon}$ and
$\mathsf{p_{min}}$, the distance metric used ($\mathsf{d_{||}}$)---necessarily
related to $\mathsf{\epsilon}$---is also considerably relevant for the final
results.

The $V$-measure \cite{rosenberg2007v} is used in this experiment to assess the
clustering results. In its simplest form, this measure is an average of
homogeneity and completeness, as shown in Eq.~\ref{eq:vm}:

\begin{equation}
    V = 2\frac{h \cdot c}{h+c}
    \label{eq:vm}
\end{equation}

\noindent Homogeneity, $h$, assesses the similarity of the observations in each
cluster, i.e., it is maximized when clusters only have observations of a single
class. Completeness, $c$, evaluates the degree to which observations from a
given class are grouped together. Therefore, it is maximized when all
observations of any given class are placed in the same cluster. The $V$-measure
provides a score between 0 and 1 that intuitively quantifies the clustering
quality, requiring maximum homogeneity and completeness---i.e., that both
evaluate to 1---to yield the perfect score of 1. Further, the metric is
independent of the number of classes, clusters, and data points, and crucially,
of the clustering algorithm used. Although other metrics exist
\cite{rosenberg2007v,palacio2019evaluation,meilua2003comparing,wagner2007comparing},
the advantages offered by the $V$-measure make it a clear and intuitive
choice for this example.

\subsection{Experimental Setup}
\label{sec:useexample:expsetup}

To keep the example simple and easily interpretable, we parameterized
\textit{Clugen} for generating 2D clusters with the mandatory parameters
specified in Table~\ref{tab:useexample:clugen}, leaving the optional parameters
to their defaults. As can be observed in this table, $l$ (\texttt{llength})
varies from 0 to 18 with increments of 6. For each value of $l$, 30 cluster sets
were generated with \textit{Clugen} using different predefined seeds.

\begin{table}[]
    \caption{Mandatory parameters used in \textit{Clugen} for testing five
    different clustering algorithms. The optional parameters are set to their
    defaults. All parameters are fixed, except $l$ (\texttt{llength}), which
    varies from 0 to 18, with of step of 6.}
    \label{tab:useexample:clugen}
    \begin{center}
    {\small
    \begin{tabular}{lll}
    \toprule
    \multicolumn{2}{l}{Symbol / Name} & Value \\
    \midrule
    $n$
        & \tabtt{num\_dims}
        & 2\\
    $c$
        & \tabtt{num\_clusters}
        & 4\\
    $p$
        & \tabtt{num\_points}
        & 2000\\
    $\mathbf{d}$
        & \tabtt{direction}
        & $(1, 1)$\\
    $\theta_\sigma$
        & \tabtt{angle\_disp}
        & $\pi/16$\\
    $\mathbf{s}$
        & \tabtt{cluster\_sep}
        & $(5, 5)$\\
    $l$
        & \tabtt{llength}
        & $\{0, 6, 12, 18\}$\\
    $l_\sigma$
        & \tabtt{llength\_disp}
        & $0.5$\\
    $f_\sigma$
        & \tabtt{lateral\_disp}
        & $1$\\

    \bottomrule
    \end{tabular}
    }
    \end{center}
\end{table}

The cluster sets generated with \textit{Clugen} were subject to the clustering
algorithms described in the previous section, parameterized as described in
Table~\ref{tab:useexample:algs}. More specifically, the number of clusters for
the $k$-means, fuzzy $c$-means, and $k$-medoids algorithms was set to 4---i.e.,
to the number of clusters set in \textit{Clugen}---with the maximum number of
iterations and convergence tolerance left to their defaults in the Clustering.jl
package. For fuzzy $c$-means, the fuzziness parameter, $\mathsf{m}$, was set to
values between 1.5 and 3.5, with a step of 0.5
\cite{pal1995cluster,zhou2014fuzziness}. For the algorithms that work directly
with dissimilarity measures---$k$-medoids, AHC, and DBSCAN---we experimented
with four commonly-used distance metrics in clustering, namely with the
Euclidean, squared Euclidean, Chebyshev, and Jaccard distances
\cite{irani2016clustering}. For AHC in particular, single, average, and complete
linkages (respectively, the minimum, mean, and maximum distance between
observations in a cluster pair) were tested. Finally, for DBSCAN,
$\mathsf{\epsilon}$ and $\mathsf{p_{min}}$ were empirically set to a range of
values according to the scale and number of observations in the generated data
sets, respectively. Table~\ref{tab:useexample:algs} summarizes all the
parameters tested in the different algorithms.

\begin{table}[]
    \caption{Parameters for the tested algorithms: $\mathsf{k}$ and $\mathsf{c}$
        specify the
        number of centroids/medoids (i.e., the number of clusters, $c$);
        $\mathsf{i_{max}}$ and $\mathsf{\varepsilon}$ are the maximum number of
        iterations and the algorithm convergence tolerance between iterations,
        respectively (these are set to the defaults defined in the Clustering.jl
        package); $\mathsf{m}$ is the degree of fuzziness in fuzzy $c$-means;
        $\mathsf{l_{||}}$
        denotes AHC's linkage. For DBSCAN, $\mathsf{\epsilon}$ is the radius of
        a point neighborhood, while $\mathsf{p_{min}}$ represents the minimum
        number of neighbors a point must have to be considered a density point.
        $\mathsf{d_{||}}$ is the distance metric for algorithms that work with
        dissimilarity measures ($k$-medoids, AHC and DBSCAN).
        Fixed parameters are denoted by the $=$ symbol, while variable
        parameters are specified with $\in$.}
    \label{tab:useexample:algs}
    \begin{center}
    {\small
    \begin{tabular}{lll}
    \toprule
    Algorithm & Parameters & \\
    \midrule
    $k$-means &
      \multicolumn{2}{l}{
        $\mathsf{k}=4$, $\mathsf{i_{max}}=100$, $\mathsf{\varepsilon}=\num{1e-6}$} \\
    Fuzzy $c$-means &
      \multicolumn{2}{l}{
        $\mathsf{c}=4$, $\mathsf{m} \in \{1.5, 2.0, 2.5, 3.0, 3.5\}$, $\mathsf{i_{max}}=100$, $\mathsf{\varepsilon}=\num{1e-3}$} \\
    $k$-medoids &
        $\mathsf{k}=4$, $\mathsf{i_{max}}=200$, $\mathsf{\varepsilon}=\num{1e-8}$ &
        \multirow{3}{*}{
        \begin{tabular}{;{1pt/1pt}p{3.1cm}}
            \footnotesize
            $\begin{aligned}
                \mathsf{d_{||}} \in \{
                    & \textrm{Euclidean},\\[-2mm]
                    & \textrm{Sq. Euclidean},\\[-2mm]
                    & \textrm{Chebyshev},\\[-2mm]
                    & \textrm{Jaccard}\}
            \end{aligned}$
        \end{tabular}
        }
        \\
    AHC &
      $\mathsf{l_{||}} \in \{\textrm{single},\textrm{average},\textrm{complete}\}$ & \\
    DBSCAN &
      $\mathsf{\epsilon} \in \{0.5, 1, 2, 4\}$, $\mathsf{p_{min}} \in \{3,5,20,50\}$ & \\
    \bottomrule
    \end{tabular}
    }
    \end{center}
\end{table}

Each of the 120 datasets generated with \textit{Clugen} (4 different values
of $l$ combined with 30 different seeds), was subject to the
aforementioned clustering algorithms and their various parameterizations, in a
total of 86 algorithm/parameter combinations, yielding \num{10320} clusterings
assessed with the $V$-measure. In effect, this is a basic parameter sweep, which
nonetheless allows this example to remain as simple as possible while
illustrating a potential use of \textit{Clugen}.

\subsection{Results}
\label{sec:useexample:results}

\begin{figure}[]
    \centering

    \subfloat[$l=0$.\label{fig:test_clustalgs:ll0}]{
        \includegraphics[width=0.5\linewidth]{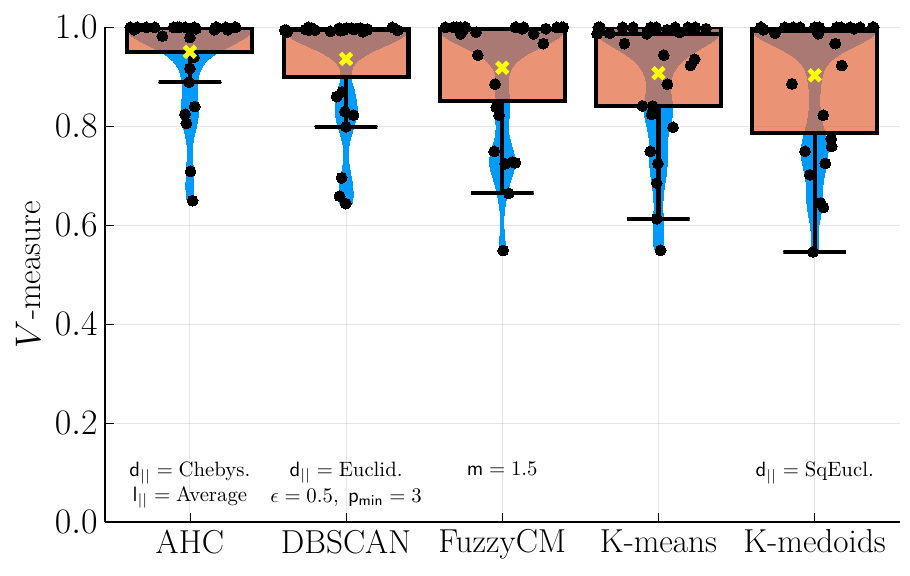}}
    \subfloat[$l=6$.\label{fig:test_clustalgs:ll6}]{
        \includegraphics[width=0.5\linewidth]{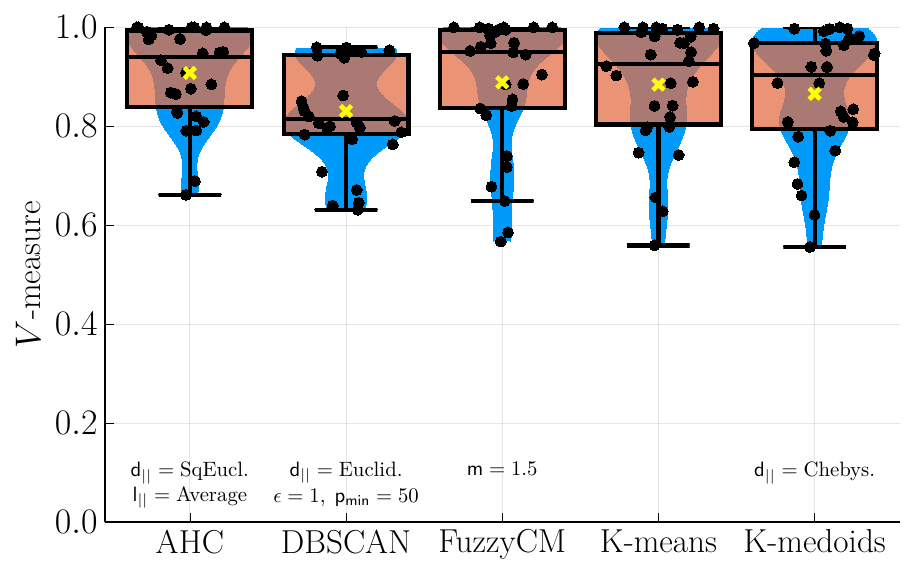}}

    \subfloat[$l=12$.\label{fig:test_clustalgs:ll12}]{
        \includegraphics[width=0.5\linewidth]{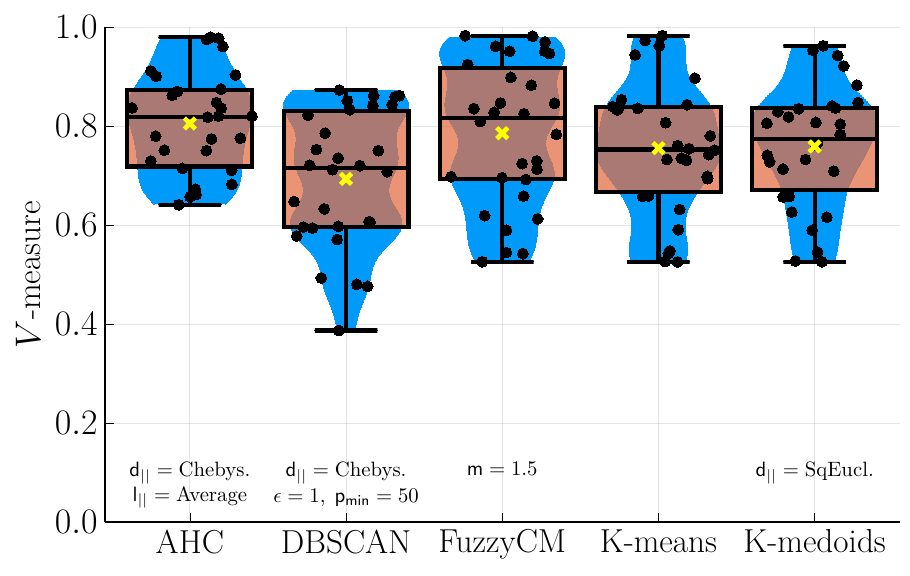}}
    \subfloat[$l=18$.\label{fig:test_clustalgs:ll18}]{
        \includegraphics[width=0.5\linewidth]{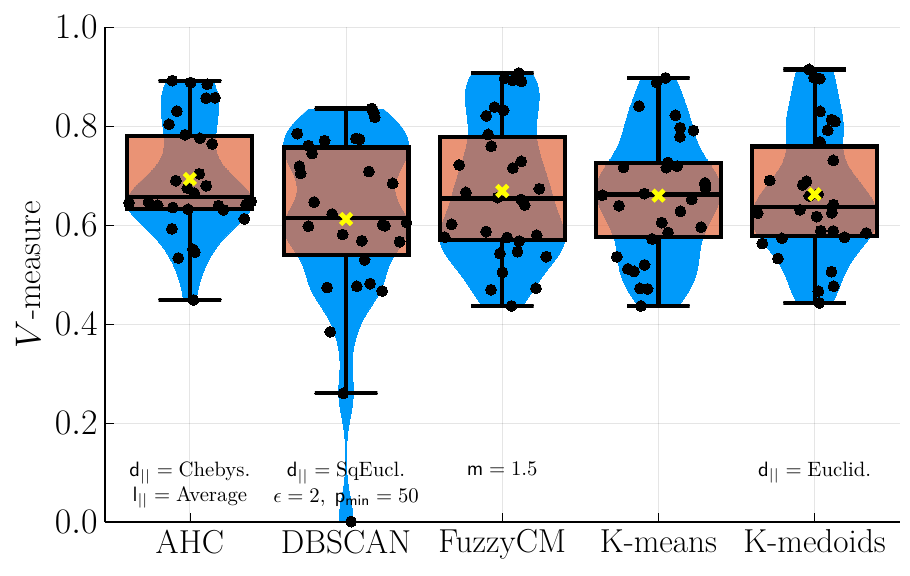}}

    \caption{$V$-measure distributions for the 5 clustering algorithms applied
        on 120 cluster sets (30 per value of $l$) generated
        with \textit{Clugen} with the parameters specified in
        Table~\ref{tab:useexample:clugen}. The mean is shown as a yellow
        \texttt{x} mark. Black dots indicate the $V$-measure obtained by
        applying the corresponding algorithm to individual cluster sets, and
        their horizontal position is randomized for visualization purposes.
        Results are shown for algorithm parameter combinations which produced
        the highest mean $V$-measure, with the corresponding parameters shown
        above the $x$-axis, where $\mathsf{d_{||}}$ is the distance metric used,
        $\mathsf{l_{||}}$ is the linkage in AHC, $\mathsf{\epsilon}$ and
        $\mathsf{p_{min}}$ correspond to DBSCAN's point neighborhood radius and
        minimum number of neighboring points, respectively, while $\mathsf{m}$
        represents the fuzziness in fuzzy $c$-means. Each subplot corresponds to
        a different value of $l$ in \textit{Clugen}, namely:
        (a) $l=0$; (b) $l=6$; (c) $l=12$; and, (d) $l=18$.}
    \label{fig:test_clustalgs}
\end{figure}

Fig.~\ref{fig:test_clustalgs} shows the $V$-measure distribution for the
parameter combination of each algorithm that yields the highest mean
$V$-measure. Each subfigure corresponds to an increasing line length, as defined
by the $l$ parameter in \textit{Clugen}.

A noticeable pattern in the results is that clustering quality, as assessed by
the $V$-measure, suffers with increasingly elongated clusters, independently of
the clustering algorithm. This is to be expected, especially in algorithms such
as $k$-means, fuzzy $c$-means, and $k$-medoids, which base their mode of
operation on the notion of clusters having an iteratively adjustable center.
Perhaps due to this reason, the AHC algorithm exhibited superior
performance compared to the other algorithms, displaying higher averages (means
and medians) and less disperse distributions of $V$-measures. The average
linkage, i.e., the average distance between all pairs of data points in any two
clusters, was the most effective for all values of $l$. AHC also tended to
prefer the Chebyshev distance---i.e., where the distance between two points is
the largest difference along coordinate dimensions.

The fuzzy $c$-means algorithm also fared well, especially for larger line
lengths. Interestingly, the algorithm produced the highest $V$-measure means
with a fuzziness value of $\mathsf{m}=1.5$, which is somewhat outside its
typical range of values according to the literature
\cite{pal1995cluster,zhou2014fuzziness}. For this reason, we performed an extra
analysis on the impact of $\mathsf{m}$ on the clustering results (available in
the supplementary materials), and observed that, for the tested values, it seems
to have little influence, perhaps except for $\mathsf{m}=3.5$, for which the
$V$-measure distribution reflects a slight decrease in clustering quality.

Conversely, DBSCAN did not adapt well to increasing line lengths, even though it
was the algorithm tested with more parameter combinations (a total of 64), and
it was the only algorithm with a $V$-measure of zero on one of the cluster sets
for $l=18$, as shown in Fig.~\ref{fig:test_clustalgs:ll18}. In turn, $k$-means
and $k$-medoids presented consistent results, generally below AHC and fuzzy
$c$-means, but clearly better than DBSCAN for increasingly elongated clusters.
Interestingly, $k$-means offered such consistency being the only algorithm for
which there was no parameter sweep, i.e., it only ran once per cluster set.

\subsection{Limitations}
\label{sec:useexample:limitations}

Given its goal of illustrating a possible use of \textit{Clugen}, the experiment
presented in this section has considerable limitations. First, results are
discussed in a descriptive fashion, with only shallow attempts at explaining
them---for example, no discussion is performed on why DBSCAN fared objectively
worse than the remaining algorithms.

Second, apart from a simple analysis of how the fuzziness parameter affects
clustering results for fuzzy $c$-means, included in the supplementary material,
no additional inquiries were performed on how $V$-measure distribution changes
with algorithm parameters. It would be particularly interesting, for example, to
further study how the different distance metrics influence clustering quality
for increasing values of $l$ for each algorithm, or even across algorithms.

Third, the implementations of $k$-means, fuzzy $c$-means, and $k$-medoids
provided by the Clustering.jl package are stochastic, meaning that they can
produce different clusterings each time they are applied to the same data.
Therefore, a more complete picture of how these algorithms behave would likely
require several runs for each of the generated datasets.

Finally, the experiment was limited to a fixed number of 2D observations, while
changing the clusters' elongation. Given the rich parameterization possible with
\textit{Clugen}, the experiment barely explores its potential.

\section{Discussion}
\label{sec:disc}

As stated in the Introduction, synthetic data generation techniques facilitate
the methodical analysis of clustering algorithms and other clustering-related
methods, which are generally influenced by the number of observations, cluster
separation, cluster shape, and other cluster features \cite{qiu2006generation}.
In particular, \textit{Clugen} allows the parameterization of these and many
other cluster characteristics, while supporting massive data generation in a
reproducible fashion, making it appropriate for experiments ranging from the
simple example presented in Section~\ref{sec:useexample}, to factorial
experiment designs \cite{qiu2006generation}, or even parameter search and
optimization tasks \cite{bergstra2012random}. For example, \textit{Clugen}
could be integrated with EA-based clustering difficulty scenarios
\cite{shand2021hawks,shand2019evolving}, or even plugged into a genetic
programming pipeline, given it accepts functions as parameters.

To further highlight \textit{Clugen}'s potential, the impact of its precursor,
\textit{generateData} \cite{fachada2020generatedata}, can be analyzed. The
\textit{generateData} algorithm, available for MATLAB/Octave, was originally
created for augmenting spectrometric data processed with principal component
analysis; the purpose was to test novel, context-specific criteria for AHC
\cite{fachada2014spectrometric}. It has since been used for evaluating
clustering algorithms and their initialization strategies
\cite{nabatian2023adaptive,molina2018estudio,molina2018d3cas,alabdulatif2018privacy,alabdulatif2019secure,alabdulatif2020towards,olukanmi2019rethinking,olukanmi2019kmeans,olukanmi2020k,olukanmi2022k},
assessing outlier detection algorithms \cite{mayanglambam2023pso}, testing the
accuracy of a method for estimating the number of clusters \cite{hao2019video},
generating data for wireless sensor networks \cite{mohammed2022sectored} and
population dynamics models \cite{zamberletti2018connectivity}, as well as
generating data for classification purposes in federated learning models
\cite{berghout2022heterogeneous}. Given that the functionality offered by
\textit{generateData} consists of a very limited subset of
MOCluGen---\textit{Clugen}'s MATLAB/Octave implementation---we believe that
\textit{Clugen} in general, and its four implementations in particular, have
tremendous potential as synthetic data generators for a variety of scientific
studies including, but not limited to, systematic analyses of clustering
techniques.

Nonetheless, \textit{Clugen} presents several limitations, mostly in the form
of missing convenience features available in some of the works discussed in
Section~\ref{sec:background}. For example:

\begin{itemize}
    \item Formal cluster separation and/or overlap guarantees are not directly
        offered. However, by manipulating the \texttt{cluster\_sep}
        parameter---the average cluster center separation---and the appropriate
        distributional properties, sufficiently large cluster set samples allow
        a desired separation/overlap tendency to show through. This is arguably
        more in line with \textit{Clugen}'s philosophy of massive generation of
        data sets with similar characteristics, as opposed to generating
        carefully crafted synthetic clusters. In any case, since the user can
        specify absolute cluster centers via the \texttt{clucenters\_fn}
        parameter, precise cluster design is possible, though achieving a
        concrete measure of separation/overlap is the user's responsibility.
    \item \textit{Clugen} does not provide explicit noise and/or outlier
        generation. However, the \texttt{clumerge()} function, offered in the
        four software implementations, provides a simple way of merging
        arbitrary cluster data, including, for example, noise and outliers.
    \item The univariate normal distribution is used by default for
        generating the distance of points from their projections, with the
        direction (from projection to point) being either completely random
        (when $p_\text{final}()$ is set to \texttt{"n"}), or random but
        orthogonal to the cluster-supporting line (when $p_\text{final}()$
        is set to \texttt{"n-1"}). It is possible to use multivariate
        distributions or different distributions for each dimension, but
        this requires the user to implement a custom $p_\text{final}()$
        function.
    \item There is no randomization of input parameters. This
        must be explicitly set up by the user, as done for the experiment
        presented in Section~\ref{sec:useexample}.
    \item \textit{Clugen} does not offer a user-friendly ``difficulty
        level''. The definition of difficulty is left entirely to the user
        through the various parameters described in Table~\ref{tab:paramsmand}
        and Table~\ref{tab:paramsopt}.
\end{itemize}

These missing features are by design since they either go against
\textit{Clugen}'s philosophy and/or hide complexity we aim to expose.
Our goal is to allow users to have complete control---yet fully
understand---an already extensive parameter set. These limitations are
essentially solvable in one way or another. However, one limitation which
might be difficult, or even impossible, to work around stems from the main
premise on which the data is generated, i.e., by using lines to support
clusters. This effectively limits the type of data which can be produced, or
force users to delve deeper into \textit{Clugen}'s code to obtain the
desired results. For example, if clusters are to be supported by arbitrary
curves, a generator such as AC \cite{li2022ac}---which natively supports Bezier
curves---will probably be easier to use.

\section{Conclusions}
\label{sec:conclusions}

In this paper, we proposed \textit{Clugen}, an algorithm and toolbox for
generating multidimensional clusters supported by line segments
using arbitrary statistical distributions. It was shown that
\textit{Clugen}, available as a fully open source, tested, and documented
solution for Python, R, Julia, and MATLAB/Octave, is capable of creating
expressive and comprehensive data, and can be integrated with experiments for
evaluating clustering techniques. Although our proposal presents some
limitations---mostly by design and highlighted in Section~\ref{sec:disc}---we
believe that, given the impact of its much more limited precursor,
\textit{generateData}, it holds the potential to be an essential component
in many research undertakings.

\section*{Supplementary Materials}

All the images presented in this paper, as well as the results presented in the
usage example in Section~\ref{sec:useexample}, were generated with CluGen.jl and
plotted with Plots.jl \cite{christ2022plots}, with the respective notebooks
available at Zenodo \cite{fachada2022notebooks}.

\section*{Acknowledgments}

This work is supported by Fundação para a Ciência e a Tecnologia under
Grant UIDB/04111/2020 (COPELABS).
The authors would also like to thank the anonymous referees for their valuable
comments and helpful suggestions.

\bibliographystyle{elsarticle-num}

\begin{thebibliography}{10}
\expandafter\ifx\csname url\endcsname\relax
  \def\url#1{\texttt{#1}}\fi
\expandafter\ifx\csname urlprefix\endcsname\relax\def\urlprefix{URL }\fi
\expandafter\ifx\csname href\endcsname\relax
  \def\href#1#2{#2} \def\path#1{#1}\fi

\bibitem{li2022ac}
W.~Li, Z.~Zhou, {AC}: A data generator for evaluation of clustering, TechRxiv
  (Feb. 2022).
\newblock \href {https://doi.org/10.36227/techrxiv.19091330.v1}
  {\path{doi:10.36227/techrxiv.19091330.v1}}.

\bibitem{korzeniewski2014empirical}
J.~Korzeniewski, Empirical evaluation of {OCLUS} and {GenRandomClust}
  algorithms of generating cluster structures, Statistics in Transition new
  series 15~(3) (2014) 487--494.

\bibitem{shand2021hawks}
C.~Shand, R.~Allmendinger, J.~Handl, A.~Webb, J.~Keane, {HAWKS}: {E}volving
  challenging benchmark sets for cluster analysis, IEEE Transactions on
  Evolutionary Computation (2021).
\newblock \href {https://doi.org/10.1109/TEVC.2021.3137369}
  {\path{doi:10.1109/TEVC.2021.3137369}}.

\bibitem{smith2015generating}
K.~Smith-Miles, S.~Bowly, Generating new test instances by evolving in instance
  space, Computers \& Operations Research 63 (2015) 102--113.
\newblock \href {https://doi.org/10.1016/j.cor.2015.04.022}
  {\path{doi:10.1016/j.cor.2015.04.022}}.

\bibitem{pei2006synthetic}
Y.~Pei, O.~Za{\"\i}ane, A synthetic data generator for clustering and outlier
  analysis, Tech. Rep. TR06-15, Computing Science Department, University of
  Alberta, Edmonton, Canada (2006).
\newblock \href {https://doi.org/10.7939/R3B23S} {\path{doi:10.7939/R3B23S}}.

\bibitem{fachada2020generatedata}
N.~Fachada, A.~C. Rosa, {generateData}--a {2D} data generator, Software Impacts
  4 (2020) 100017.
\newblock \href {https://doi.org/10.1016/j.simpa.2020.100017}
  {\path{doi:10.1016/j.simpa.2020.100017}}.

\bibitem{shand2019evolving}
C.~Shand, R.~Allmendinger, J.~Handl, A.~Webb, J.~Keane, Evolving controllably
  difficult datasets for clustering, in: Proceedings of the Genetic and
  Evolutionary Computation Conference, GECCO '19, ACM, New York, NY, USA, 2019,
  pp. 463--471.
\newblock \href {https://doi.org/10.1145/3321707.3321761}
  {\path{doi:10.1145/3321707.3321761}}.

\bibitem{macia2014towards}
N.~Macia, E.~Bernad{\'o}-Mansilla, Towards {UCI}+: a mindful repository design,
  Information Sciences 261 (2014) 237--262.
\newblock \href {https://doi.org/10.1016/j.ins.2013.08.059}
  {\path{doi:10.1016/j.ins.2013.08.059}}.

\bibitem{iglesias2019mdcgen}
F.~Iglesias, T.~Zseby, D.~Ferreira, A.~Zimek, {MDCGen}: Multidimensional
  dataset generator for clustering, Journal of Classification 36~(3) (2019)
  599--618.
\newblock \href {https://doi.org/10.1007/s00357-019-9312-3}
  {\path{doi:10.1007/s00357-019-9312-3}}.

\bibitem{qiu2006generation}
W.~Qiu, H.~Joe, Generation of random clusters with specified degree of
  separation, Journal of Classification 23~(2) (2006) 315--334.
\newblock \href {https://doi.org/10.1007/s00357-006-0018-y}
  {\path{doi:10.1007/s00357-006-0018-y}}.

\bibitem{melnykov2012mixsim}
V.~Melnykov, W.-C. Chen, R.~Maitra, {MixSim}: An {R} package for simulating
  data to study performance of clustering algorithms, Journal of Statistical
  Software 51~(12) (2012) 1--25.
\newblock \href {https://doi.org/10.18637/jss.v051.i12}
  {\path{doi:10.18637/jss.v051.i12}}.

\bibitem{handl2006multi}
J.~Handl, J.~Knowles, Multi-objective clustering and cluster validation,
  Springer, 2006, Ch.~2, pp. 21--47.
\newblock \href {https://doi.org/10.1007/3-540-33019-4_2}
  {\path{doi:10.1007/3-540-33019-4_2}}.

\bibitem{steinley2005oclus}
D.~Steinley, R.~Henson, {OCLUS}: an analytic method for generating clusters
  with known overlap, Journal of classification 22~(2) (2005) 221--250.
\newblock \href {https://doi.org/10.1007/s00357-005-0015-6}
  {\path{doi:10.1007/s00357-005-0015-6}}.

\bibitem{vanrossum2022python}
G.~Van~Rossum, F.~L. Drake, Python 3 Reference Manual, CreateSpace, Scotts
  Valley, CA, 2009.

\bibitem{r2022}
{R Core Team}, \href{https://www.R-project.org/}{R: A Language and Environment
  for Statistical Computing}, R Foundation for Statistical Computing, Vienna,
  Austria (2022).
\newline\urlprefix\url{https://www.R-project.org/}

\bibitem{bezanson2017julia}
J.~Bezanson, A.~Edelman, S.~Karpinski, V.~B. Shah, Julia: A fresh approach to
  numerical computing, SIAM Review 59~(1) (2017) 65--98.
\newblock \href {https://doi.org/10.1137/141000671}
  {\path{doi:10.1137/141000671}}.

\bibitem{matlab2022}
{The MathWorks Inc.},
  \href{https://www.mathworks.com/products/matlab.html}{MATLAB}, Natick,
  Massachusetts (2022).
\newline\urlprefix\url{https://www.mathworks.com/products/matlab.html}

\bibitem{octave2022}
J.~W. Eaton, D.~Bateman, S.~Hauberg, R.~Wehbring,
  \href{https://www.gnu.org/software/octave/}{{GNU Octave} manual: a high-level
  interactive language for numerical computations} (2022).
\newline\urlprefix\url{https://www.gnu.org/software/octave/}

\bibitem{handl2005cluster}
J.~Handl, J.~Knowles, Cluster generators for large high-dimensional data sets
  with large numbers of clusters (2005).

\bibitem{milligan1985algorithm}
G.~W. Milligan, An algorithm for generating artificial test clusters,
  Psychometrika 50~(1) (1985) 123--127.
\newblock \href {https://doi.org/10.1007/BF02294153}
  {\path{doi:10.1007/BF02294153}}.

\bibitem{pape2000clusutils}
D.~X. Pape, D.~S. Dubin, \href{http://clusutils.sourceforge.net/}{Clusutils},
  SourceForge (2000).
\newline\urlprefix\url{http://clusutils.sourceforge.net/}

\bibitem{vennam2005syndeca}
J.~R. Vennam, S.~Vadapalli, {SynDECA}: A tool to generate synthetic datasets
  for evaluation of clustering algorithms, in: J.~R. Haritsa, T.~M. Vijayaraman
  (Eds.), Advances in Data Management 2005: Proceedings of the Eleventh
  International Conference on Management of Data, COMAD 2005, Computer Society
  of India, 2005, pp. 27--36.

\bibitem{pedregosa2011scikit}
F.~Pedregosa, G.~Varoquaux, A.~Gramfort, V.~Michel, B.~Thirion, O.~Grisel,
  M.~Blondel, P.~Prettenhofer, R.~Weiss, V.~Dubourg, J.~Vanderplas, A.~Passos,
  D.~Cournapeau, M.~Brucher, M.~Perrot, {{\'E}}douard Duchesnay,
  \href{http://jmlr.org/papers/v12/pedregosa11a.html}{Scikit-learn: Machine
  learning in {P}ython}, Journal of Machine Learning Research 12~(85) (2011)
  2825--2830.
\newline\urlprefix\url{http://jmlr.org/papers/v12/pedregosa11a.html}

\bibitem{maitra2010simulating}
R.~Maitra, V.~Melnykov, Simulating data to study performance of finite mixture
  modeling and clustering algorithms, Journal of Computational and Graphical
  Statistics 19~(2) (2010) 354--376.
\newblock \href {https://doi.org/10.1198/jcgs.2009.08054}
  {\path{doi:10.1198/jcgs.2009.08054}}.

\bibitem{schubert2015framework}
E.~Schubert, A.~Koos, T.~Emrich, A.~Z{\"u}fle, K.~A. Schmid, A.~Zimek, A
  framework for clustering uncertain data, Proceedings of the VLDB Endowment
  8~(12) (2015) 1976--1979.
\newblock \href {https://doi.org/10.14778/2824032.2824115}
  {\path{doi:10.14778/2824032.2824115}}.

\bibitem{rousseeuw1987silhouettes}
P.~J. Rousseeuw, Silhouettes: a graphical aid to the interpretation and
  validation of cluster analysis, Journal of Computational and Applied
  Mathematics 20 (1987) 53--65.
\newblock \href {https://doi.org/10.1016/0377-0427(87)90125-7}
  {\path{doi:10.1016/0377-0427(87)90125-7}}.

\bibitem{fachada2014spectrometric}
N.~Fachada, M.~A.~T. Figueiredo, V.~V. Lopes, R.~C. Martins, A.~C. Rosa,
  Spectrometric differentiation of yeast strains using minimum volume increase
  and minimum direction change clustering criteria, Pattern Recognit. Lett. 45
  (2014) 55--61.
\newblock \href {https://doi.org/10.1016/j.patrec.2014.03.008}
  {\path{doi:10.1016/j.patrec.2014.03.008}}.

\bibitem{zellinger2023repliclust}
M.~J. Zellinger, P.~B{\"u}hlmann, repliclust: Synthetic data for cluster
  analysis, arXiv, 2303.14301 (2023).
\newblock \href {https://doi.org/10.48550/arXiv.2303.14301}
  {\path{doi:10.48550/arXiv.2303.14301}}.

\bibitem{harris2020numpy}
C.~R. Harris, K.~J. Millman, S.~J. van~der Walt, R.~Gommers, P.~Virtanen,
  D.~Cournapeau, E.~Wieser, J.~Taylor, S.~Berg, N.~J. Smith, R.~Kern, M.~Picus,
  S.~Hoyer, M.~H. van Kerkwijk, M.~Brett, A.~Haldane, J.~Fernández~del Río,
  M.~Wiebe, P.~Peterson, P.~Gérard-Marchant, K.~Sheppard, T.~Reddy,
  W.~Weckesser, H.~Abbasi, C.~Gohlke, T.~E. Oliphant, Array programming with
  {NumPy}, Nature 585 (2020) 357--362.
\newblock \href {https://doi.org/10.1038/s41586-020-2649-2}
  {\path{doi:10.1038/s41586-020-2649-2}}.

\bibitem{viechtbauer2022mathjaxr}
W.~Viechtbauer, \href{https://CRAN.R-project.org/package=mathjaxr}{mathjaxr:
  Using 'Mathjax' in Rd Files}, {R package version 1.6-0} (2022).
\newline\urlprefix\url{https://CRAN.R-project.org/package=mathjaxr}

\bibitem{clustering2012jl}
J.~M. White, D.~Lin, A.~Stukalov, et~al.,
  \href{https://github.com/JuliaStats/Clustering.jl}{Clustering.jl -- a {J}ulia
  package for data clustering}, GitHub, accessed 11/11/2022, updated 23/09/2022
  (2012).
\newline\urlprefix\url{https://github.com/JuliaStats/Clustering.jl}

\bibitem{fachada2022notebooks}
N.~Fachada, \href{https://doi.org/10.5281/zenodo.7566684}{Supplementary
  materials for ``generating multidimensional clusters with support lines''},
  Zenodo, accessed 24/01/2023, updated 24/01/2023 (2023).
\newblock \href {https://doi.org/10.5281/zenodo.7566684}
  {\path{doi:10.5281/zenodo.7566684}}.
\newline\urlprefix\url{https://doi.org/10.5281/zenodo.7566684}

\bibitem{hastie2009elements}
T.~Hastie, R.~Tibshirani, J.~Friedman, The Elements of Statistical Learning:
  Data Mining, Inference, and Prediction, 2nd Edition, Springer Series in
  Statistics, Springer, 2009.
\newblock \href {https://doi.org/10.1007/978-0-387-84858-7}
  {\path{doi:10.1007/978-0-387-84858-7}}.

\bibitem{lloyd1982least}
S.~Lloyd, Least squares quantization in {PCM}, IEEE Transactions on Information
  Theory 28~(2) (1982) 129--137.
\newblock \href {https://doi.org/10.1109/TIT.1982.1056489}
  {\path{doi:10.1109/TIT.1982.1056489}}.

\bibitem{arthur2006k}
D.~Arthur, S.~Vassilvitskii, K-means++: The advantages of careful seeding, in:
  Proceedings of the Eighteenth Annual ACM-SIAM Symposium on Discrete
  Algorithms, SODA '07, SIAM, USA, 2007, pp. 1027--1035.

\bibitem{dunn1973fuzzy}
J.~C. Dunn, A fuzzy relative of the {ISODATA} process and its use in detecting
  compact well-separated clusters, Journal of Cybernetics 3~(3) (1973) 32--57.
\newblock \href {https://doi.org/10.1080/01969727308546046}
  {\path{doi:10.1080/01969727308546046}}.

\bibitem{schwammle2010simple}
V.~Schw{\"a}mmle, O.~N. Jensen, A simple and fast method to determine the
  parameters for fuzzy c--means cluster analysis, Bioinformatics 26~(22) (2010)
  2841--2848.
\newblock \href {https://doi.org/10.1093/bioinformatics/btq534}
  {\path{doi:10.1093/bioinformatics/btq534}}.

\bibitem{schubert2019faster}
E.~Schubert, P.~J. Rousseeuw, Faster $k$-medoids clustering: improving the
  {PAM}, {CLARA}, and {CLARANS} algorithms, in: G.~Amato, C.~Gennaro, V.~Oria,
  M.~Radovanovi{\'{c}} (Eds.), Similarity Search and Applications (SISAP 2019),
  Vol. 11807 of Lecture Notes in Computer Science, Springer International
  Publishing, 2019, pp. 171--187.
\newblock \href {https://doi.org/10.1007/978-3-030-32047-8_16}
  {\path{doi:10.1007/978-3-030-32047-8_16}}.

\bibitem{kaufman1990partitioning}
L.~Kaufman, P.~J. Rousseeuw, Partitioning around medoids ({P}rogram {PAM}), in:
  Finding Groups in Data: An Introduction to Cluster Analysis, Vol. 344, John
  Wiley \& Sons, Ltd, 1990, Ch.~2, pp. 68--125.
\newblock \href {https://doi.org/10.1002/9780470316801.ch2}
  {\path{doi:10.1002/9780470316801.ch2}}.

\bibitem{ester1996density}
M.~Ester, H.-P. Kriegel, J.~Sander, X.~Xu, A density-based algorithm for
  discovering clusters in large spatial databases with noise, in: E.~Simoudis,
  J.~Han, U.~M. Fayyad (Eds.), Proceedings of the Second International
  Conference on Knowledge Discovery and Data Mining, KDD-96, AAAI Press, 1996,
  pp. 226--231.

\bibitem{rosenberg2007v}
A.~Rosenberg, J.~Hirschberg, V-measure: A conditional entropy-based external
  cluster evaluation measure, in: J.~Eisner (Ed.), Proceedings of the 2007
  Joint Conference on Empirical Methods in Natural Language Processing and
  Computational Natural Language Learning, EMNLP-CoNLL 2007, The Association
  for Computational Linguistics, Stroudsburg, PA, USA, 2007, pp. 410--420.
\newblock \href {https://doi.org/10.7916/D80V8N84}
  {\path{doi:10.7916/D80V8N84}}.

\bibitem{palacio2019evaluation}
J.-O. Palacio-Ni{\~n}o, F.~Berzal,
  \href{https://arxiv.org/abs/1905.05667}{Evaluation metrics for unsupervised
  learning algorithms}, arXiv:1905.05667 (2019).
\newblock \href {https://doi.org/10.48550/ARXIV.1905.05667}
  {\path{doi:10.48550/ARXIV.1905.05667}}.
\newline\urlprefix\url{https://arxiv.org/abs/1905.05667}

\bibitem{meilua2003comparing}
M.~Meil{\u{a}}, Comparing clusterings by the variation of information, in:
  B.~Sch{\"o}lkopf, M.~K. Warmuth (Eds.), Learning Theory and Kernel Machines,
  Springer, 2003, pp. 173--187.
\newblock \href {https://doi.org/10.1007/978-3-540-45167-9_14}
  {\path{doi:10.1007/978-3-540-45167-9_14}}.

\bibitem{wagner2007comparing}
S.~Wagner, D.~Wagner, Comparing clusterings: an overview (2007).

\bibitem{pal1995cluster}
N.~R. Pal, J.~C. Bezdek, On cluster validity for the fuzzy c-means model, IEEE
  Transactions on Fuzzy systems 3~(3) (1995) 370--379.
\newblock \href {https://doi.org/10.1109/91.413225}
  {\path{doi:10.1109/91.413225}}.

\bibitem{zhou2014fuzziness}
K.~Zhou, C.~Fu, S.~Yang, Fuzziness parameter selection in fuzzy c-means: the
  perspective of cluster validation, Science China Information Sciences 57~(11)
  (2014) 1--8.
\newblock \href {https://doi.org/10.1007/s11432-014-5146-0}
  {\path{doi:10.1007/s11432-014-5146-0}}.

\bibitem{irani2016clustering}
J.~Irani, N.~Pise, M.~Phatak, Clustering techniques and the similarity measures
  used in clustering: a survey, International Journal of Computer Applications
  134~(7) (2016) 9--14.
\newblock \href {https://doi.org/10.5120/ijca2016907841}
  {\path{doi:10.5120/ijca2016907841}}.

\bibitem{bergstra2012random}
J.~Bergstra, Y.~Bengio,
  \href{http://jmlr.org/papers/v13/bergstra12a.html}{Random search for
  hyper-parameter optimization}, Journal of Machine Learning Research 13~(10)
  (2012) 281--305.
\newline\urlprefix\url{http://jmlr.org/papers/v13/bergstra12a.html}

\bibitem{nabatian2023adaptive}
M.~Nabatian, J.~Tanha, A.~R. Ebrahimzadeh, A.~Phirouznia, An adaptive scaling
  technique to quantum clustering, International Journal of Modern Physics C
  34~(01) (2023) 2350002.
\newblock \href {https://doi.org/10.1142/S012918312350002X}
  {\path{doi:10.1142/S012918312350002X}}.

\bibitem{molina2018estudio}
R.~Molina, \href{http://sedici.unlp.edu.ar/handle/10915/82400}{Estudio e
  implementaci{\'o}n de una t{\'e}cnica de clustering din{\'a}mico para
  trabajar con flujos de datos}, Graduation thesis, Universidad Nacional de La
  Plata, La Plata, Argentina (Jul. 2018).
\newline\urlprefix\url{http://sedici.unlp.edu.ar/handle/10915/82400}

\bibitem{molina2018d3cas}
R.~Molina, W.~Hasperu{\'e},
  \href{http://sedici.unlp.edu.ar/handle/10915/73223}{{D3CAS}: un algoritmo de
  clustering para el procesamiento de flujos de datos en {S}park}, in: XXIV
  Congreso Argentino de Ciencias de la Computaci{\'o}n, CACIC '18, 2018, pp.
  452--461.
\newline\urlprefix\url{http://sedici.unlp.edu.ar/handle/10915/73223}

\bibitem{alabdulatif2018privacy}
A.~Alabdulatif,
  \href{https://researchbank.rmit.edu.au/view/rmit:162567}{Privacy-preserving
  data analytics in cloud computing}, Ph.D. thesis, RMIT University, Melbourne,
  Australia (Nov. 2018).
\newline\urlprefix\url{https://researchbank.rmit.edu.au/view/rmit:162567}

\bibitem{alabdulatif2019secure}
A.~Alabdulatif, I.~Khalil, X.~Yi, M.~Guizani, Secure edge of things for smart
  healthcare surveillance framework, IEEE Access 7 (2019) 31010--31021.
\newblock \href {https://doi.org/10.1109/ACCESS.2019.2899323}
  {\path{doi:10.1109/ACCESS.2019.2899323}}.

\bibitem{alabdulatif2020towards}
A.~Alabdulatif, I.~Khalil, X.~Yi,
  \href{http://www.sciencedirect.com/science/article/pii/S0743731519300887}{Towards
  secure big data analytic for cloud-enabled applications with fully
  homomorphic encryption}, J. Parallel Distrib. Comput. 137 (2020) 192--204.
\newblock \href {https://doi.org/10.1016/j.jpdc.2019.10.008}
  {\path{doi:10.1016/j.jpdc.2019.10.008}}.
\newline\urlprefix\url{http://www.sciencedirect.com/science/article/pii/S0743731519300887}

\bibitem{olukanmi2019rethinking}
P.~Olukanmi, F.~Nelwamondo, T.~Marwala, Rethinking $k$-means clustering in the
  age of massive datasets: a constant-time approach, Neural. Comput. Appl.
  (2019) 1--23\href {https://doi.org/10.1007/s00521-019-04673-0}
  {\path{doi:10.1007/s00521-019-04673-0}}.

\bibitem{olukanmi2019kmeans}
P.~Olukanmi, F.~Nelwamondo, T.~Marwala, $k$-{M}eans-{L}ite++: The combined
  advantage of sampling and seeding, in: 2019 6th International Conference on
  Soft Computing \& Machine Intelligence, ISCMI '19, 2019, pp. 223--227.
\newblock \href {https://doi.org/10.1109/ISCMI47871.2019.9004300}
  {\path{doi:10.1109/ISCMI47871.2019.9004300}}.

\bibitem{olukanmi2020k}
P.~O. Olukanmi, F.~Nelwamondo, T.~Marwala, k-means-{MIND}: an efficient
  alternative to repetitive k-means runs, in: 2020 7th International Conference
  on Soft Computing \& Machine Intelligence (ISCMI), IEEE, 2020, pp. 172--176.
\newblock \href {https://doi.org/10.1109/ISCMI51676.2020.9311598}
  {\path{doi:10.1109/ISCMI51676.2020.9311598}}.

\bibitem{olukanmi2022k}
P.~Olukanmi, F.~Nelwamondo, T.~Marwala, k-means-{MIND}: comparing seeds without
  repeated k-means runs, Neural Computing and Applications (2022) 1--15\href
  {https://doi.org/10.1007/s00521-022-07554-1}
  {\path{doi:10.1007/s00521-022-07554-1}}.

\bibitem{mayanglambam2023pso}
S.~D. Mayanglambam, S.-J. Horng, R.~Pamula, {PSO} clustering and pruning-based
  {KNN} for outlier detection, Soft Computing (2023) 1--17\href
  {https://doi.org/10.1007/s00500-023-08718-4}
  {\path{doi:10.1007/s00500-023-08718-4}}.

\bibitem{hao2019video}
P.~Hao, E.~Manhando, T.~Ye, C.~Bai, Video summarization based on sparse
  subspace clustering with automatically estimated number of clusters, in:
  Proceedings of the ACM Multimedia Asia, MMAsia '19, ACM, New York, NY, USA,
  2019, Ch.~8.
\newblock \href {https://doi.org/10.1145/3338533.3366593}
  {\path{doi:10.1145/3338533.3366593}}.

\bibitem{mohammed2022sectored}
F.~A. Mohammed, N.~Mekky, H.~H. Suleiman, N.~A. Hikal, Sectored {LEACH}
  ({S-LEACH}): An enhanced {LEACH} for wireless sensor network, IET Wireless
  Sensor Systems 12~(2) (2022) 56--66.
\newblock \href {https://doi.org/10.1049/wss2.12036}
  {\path{doi:10.1049/wss2.12036}}.

\bibitem{zamberletti2018connectivity}
P.~Zamberletti, M.~Zaffaroni, F.~Accatino, I.~F. Creed, C.~De~Michele,
  Connectivity among wetlands matters for vulnerable amphibian populations in
  wetlandscapes, Ecol. Model. 384 (2018) 119--127.
\newblock \href {https://doi.org/10.1016/j.ecolmodel.2018.05.008}
  {\path{doi:10.1016/j.ecolmodel.2018.05.008}}.

\bibitem{berghout2022heterogeneous}
T.~Berghout, T.~Bentrcia, M.~A. Ferrag, M.~Benbouzid, A heterogeneous federated
  transfer learning approach with extreme aggregation and speed, Mathematics
  10~(19) (2022) 3528.
\newblock \href {https://doi.org/10.3390/math10193528}
  {\path{doi:10.3390/math10193528}}.

\bibitem{christ2022plots}
S.~Christ, D.~Schwabeneder, C.~Rackauckas, M.~K. Borregaard, T.~Breloff,
  Plots.jl -- a user extendable plotting {API} for the {Julia} programming
  language, Journal of Open Research Software (Feb. 2023).
\newblock \href {https://doi.org/10.5334/jors.431}
  {\path{doi:10.5334/jors.431}}.

\end{thebibliography}

\end{document}